%% file: main.tex
\documentclass{article}

\usepackage[preprint]{neurips_2026}

\usepackage[utf8]{inputenc}
\usepackage[T1]{fontenc}
\usepackage[hidelinks]{hyperref}
\usepackage{url}
\usepackage{booktabs}
\usepackage{amsfonts}
\usepackage{amsmath}
\usepackage{amssymb}
\usepackage{nicefrac}
\usepackage{microtype}
\usepackage{xcolor}
\usepackage{graphicx}
\usepackage{subcaption}
\usepackage{multirow}
\usepackage{enumitem}
\usepackage{placeins}

\graphicspath{{figures/}}

\title{Refit the Probe:\\ Single-Direction Ablation Is Not a Necessity Test}

\author{%
  Yuhang Jiang \\
  University of Trento, Italy \\
  \texttt{jyhtjtj@gmail.com}
}

\begin{document}

\maketitle

\input{sections/abstract}
\input{sections/introduction}
\input{sections/related_work}
\input{sections/setup}
\input{sections/observation}
\input{sections/tap}
\input{sections/notnecessity}
\input{sections/erasure}
\input{sections/depth}
\input{sections/survive}
\input{sections/conclusion}

\bibliographystyle{plainnat}
\bibliography{references}

\newpage
\appendix
\input{sections/appendix}

\end{document}

%% file: sections/abstract.tex
\begin{abstract}
Probes are routinely paired with an intervention: ablate the direction the probe found, run the
model, and read the change in task accuracy, taking a large drop as evidence that the computation
depends on what the probe read and a near-zero drop as evidence that it does not. Either inference requires that the ablation have removed the target from the layer. We find that the
ablation does not remove what it targets. A probe refitted on the ablated activations recovers its
original accuracy in every cell we test, and keeps recovering when the probe's entire row space is
deleted rather than a single axis, because the quantity survives in the orthogonal complement.
Because a refitted probe recovers, neither a large task drop nor a near-zero one establishes
whether the model needed the target, and one probe fit detects this. Replacing the ablation with iterative nullspace
projection, scored against random subspaces of matched dimension, reverses the conclusion:
representations that looked causally inert carry most of the task. The correction also separates
where a variable is most readable from where deleting it does most damage, and those are not the
same layer in any pretrained model we study. The erasure is defined by a linear probe family, so
removing a nonlinearly encoded quantity remains open.
\end{abstract}

%% file: sections/introduction.tex
\section{Introduction}
\label{sec:intro}

\begin{figure}[!t]
\centering
\includegraphics[width=\linewidth]{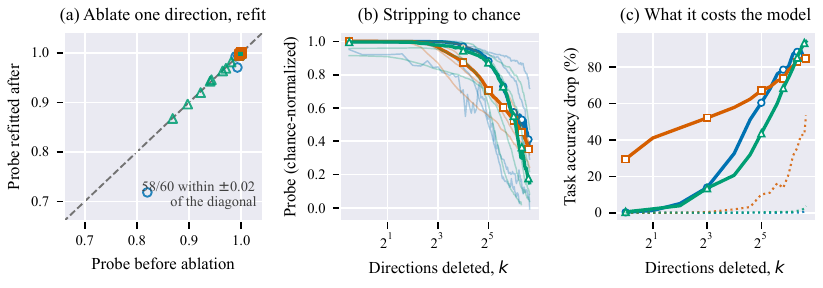}

\vspace{0.25em}
\centerline{\includegraphics[width=0.95\linewidth]{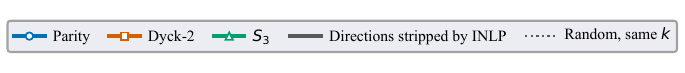}}
\caption{One measurement read at three points, over the $15$ formal-task cells at $n{=}4$ seeds. \textbf{(a)} Ablate the probe's own direction, then refit. \textbf{(b)} Probe accuracy as directions are stripped one at a time, thin lines per architecture, thick lines the task median. \textbf{(c)} What deleting those directions costs the running model, against random subspaces of the same size. Accuracy in (a) and (b) is chance-normalized; Section~\ref{sec:notnec} reads (a) against (c).}
\label{fig:headline}
\end{figure}

Probing establishes that a variable is linearly readable from a layer. Closing the gap between that and the model using it takes an intervention: delete the direction the probe found, run the model, and read the change in task accuracy. A large drop is reported as evidence that the computation depends on the direction, and a drop near zero as evidence that it does not.

The intervention does not remove the information. Refitting a probe on the ablated activations, at the same layer, recovers the original accuracy: $1.000$ where it read $1.000$ before on Dyck Transformer L0, $0.987$ to $0.988$ on Parity Mamba L3 (Fig.~\ref{fig:headline}a). It still recovers after the probe's full $K{-}1$ row space is deleted rather than a single axis. The quantity is in the orthogonal complement the whole time, so the ablation measured how much a particular linear format was disturbed, not whether a feature was needed. Both readings of the result are unavailable: the large drops and the zeros alike. The check that catches this is one probe fit, and we run it for every cell.

We then replace the measurement. Iterative nullspace projection (INLP)~\citep{ravfogel2020null} strips probe directions until no linear probe beats chance, which gives a subspace whose removal leaves the target linearly undecodable wherever the iteration reaches chance within budget. Deleting that subspace inside the running model is a necessity test, and scoring it against a random subspace of the same dimension makes the number readable, since deleting $k$ arbitrary directions is damaging on its own. The corrected measurement reverses the conclusion the single-direction version supported: the readable representations are load-bearing, costing the two pretrained models $0.71$ and $0.89$ of relative accuracy where matched random subspaces cost under $0.03$.

Amnesic probing~\citep{elazar2021amnesic} already pairs that erasure with a behavioural read-out, so the instrument is not new. We add the machinery that makes its output comparable across cells. The erasure runs across architectures, which forces the question of which tensor is being read and written (Section~\ref{sec:tap}). It is scored against random subspaces of matched dimension at a budget shared across cells: on Dyck the random control begins to climb past $k{=}32$, so a drop quoted at each cell's own $k$ is not comparable to the next cell's (Fig.~\ref{fig:inlp_curve}). The refit check is run and reported per cell rather than assumed. Readability and single-direction fragility are then separated layer by layer, and they do come apart, in the same direction in every pretrained model we ran: the probe peaks several blocks above the ablation peak. A layerwise profile is therefore not fixed by either measurement alone.

Whether any of these numbers may be read at all turns on three properties of the instrument. Probing and intervention must act on one tensor, which is not automatic across architectures: a Mamba block returns a mixer output and a residual separately, and reading one while writing the other compares two different quantities (Section~\ref{sec:tap}). The map out of standardized probe space can return a direction carrying less variance than a random one, at which point its ablation is uninterpretable. And ``the best probe layer'' is an argmax over ties wherever the probe saturates, so the measured drop follows the tie-break rather than the model (Section~\ref{sec:notnec}).

This is a paper about whether an interpretability measurement supports the claim attached to it. Section~\ref{sec:survive} reports where our own claims stop, and why the observation of Section~\ref{sec:observation} is not offered as a result. The vehicle is state tracking in small sequence models. Transformers~\citep{vaswani2017attention} aggregate prefix information through attention; state-space models such as Mamba~\citep{gu2023mamba} update a recurrent hidden state through selective linear dynamics. Section~\ref{sec:observation} reports the observation that motivated the study, that layerwise profiles move with the task rather than sitting fixed per architecture; making it predictive is one of the attempts that failed.

%% file: sections/related_work.tex
\section{Related work}
\label{sec:related}

\paragraph{State-space models and their expressive limits.}
Structured state-space models~\citep{gu2022efficiently} and selective SSMs such as Mamba~\citep{gu2023mamba} are an efficient recurrent alternative to attention, and Mamba-2~\citep{dao2024transformers} connects the two through structured state-space duality. A parallel line bounds what SSMs can represent on formal languages. They cannot express computation outside $\mathrm{TC}^0$ and cannot exactly solve permutation composition~\citep{merrill2024illusion}, but they do handle star-free state tracking and bounded hierarchical structure without simulating a stack~\citep{sarrof2024expressive}. Single-layer diagonal selective SSMs without $B$-matrix selection are restricted to commutative automata~\citep{terzic2025expressiveness}, with structured sparse transitions proposed as a remedy~\citep{terzic2025structured}. Linear RNNs with only positive eigenvalues cannot solve parity, while negative eigenvalues let Mamba and DeltaNet~\citep{yang2024deltanet} solve it~\citep{grazzi2025unlocking}, and Mamba-3~\citep{lahoti2026mamba3} takes complex-valued state updates for the same reason. \citet{bick2025gather} find a shared gather-and-aggregate mechanism behind in-context retrieval in both families, attributing the recall gap to SSMs running it with smoother attention rather than to its absence.

\paragraph{Length generalization.}
Length generalization has been analyzed from the Transformer side~\citep{anil2022exploring,li2025vanishing} and the recurrent side~\citep{ruiz2025understanding,lu2025mamba}. Positional encoding variants such as ALiBi~\citep{press2022alibi} and RoPE~\citep{su2024roformer} are Transformer-side interventions. We include ALiBi, RoPE, and a Gated Transformer as controls, and evaluate every architecture at three times the training length.

\paragraph{Formal languages as mechanistic testbeds.}
Formal languages have been used to probe what recurrent and attention-based models represent~\citep{weiss2018practical,suzgun2019lstm,bhattamishra2020ability,deletang2023neural}; \citet{strobl2024formal} survey expressivity results for Transformers and RNNs. \citet{hahn2020theoretical} proves that fixed-size self-attention cannot model periodic finite-state languages such as Parity, nor unbounded hierarchical structure such as Dyck, unless architectural resources such as layers or heads grow with input length; \citet{hewitt2020rnns} and \citet{yao2021self} give an $O(m\log k)$-memory recurrent construction and a constant-depth $O(\log k)$-memory soft-attention construction for bounded hierarchical languages. Our trained-model behavior is consistent with this picture on both counts (Appendix Figs.~\ref{fig:probes} and~\ref{fig:lengthgen}). Earlier mechanistic analyses focus on one architecture family at a time: \citet{liu2023shortcuts,liu2023flipflop} and \citet{wen2023transformers} probe Transformer on finite automata, flip-flop, and bounded Dyck respectively. None runs the within-architecture cross-task comparison of Section~\ref{sec:observation}: whether a given architecture keeps its layerwise strategy as the task changes. The reversal we see there is the motivation for this paper rather than one of its results.

\paragraph{Probing and causal intervention.}
Linear probes~\citep{alain2016understanding}, dating to subject-verb agreement analyses of LSTMs~\citep{linzen2016assessing} and systematically surveyed by~\citet{belinkov2019analysis}, and causal interventions such as activation patching~\citep{vig2020investigating,meng2022locating}, causal abstraction~\citep{geiger2021causal}, circuit discovery~\citep{wang2022interpretability} and circuit-level progress measures~\citep{nanda2023progress} are standard tools in mechanistic interpretability. We use both because readability alone can be misleading, a point that becomes especially important in pretrained models where the best probe layer and the layer with the largest ablation response can diverge. \citet{elazar2021amnesic} make this argument with a sharper instrument, erasing a property by iterative nullspace projection~\citep{ravfogel2020null} and reading the change in the model's own behavior. That instrument is theirs and Section~\ref{sec:intro} lists what we add to it. This paper argues against the more common single-direction form, where the probe direction alone is projected out. A probe refitted on the ablated layer recovers its original accuracy, so the deletion removed no information, and the near-zero drops reported from it describe the cut rather than the model.

Two results are close to ours and fail in the opposite direction. \citet{makelov2024subspace} show that a subspace intervention can move the output \emph{as if} a feature had changed while in fact routing through a dormant parallel pathway, so the intervention succeeds for the wrong reason; here it does not remove the feature at all, and so cannot succeed for any reason. \citet{hase2023localization} find that causal-tracing localization does not predict which layer is best to edit, which is the precedent for the depth comparison of Section~\ref{sec:depth} on a different pair of measurements. Concurrent cross-architecture work using SAE features finds Mamba's induction circuits structurally analogous to those in Transformers~\citep{wang2025universality}, indicating single-task mechanistic similarity; the task-axis observation here is complementary, documenting within-architecture profile reversal across tasks.

\paragraph{Component attribution in SSMs.}
\citet{arora2024zoology} analyze the gated-convolution--vs--attention gap on associative recall and show that input-dependent sparse attention can close most of it; \citet{arora2025mechanistic} then find via causal interventions that Mamba implements induction through the short convolution rather than the selective SSM. Our Conv1D ablation under a matched pure-PyTorch backend (Fig.~\ref{fig:noconv}) shows that the same component effect varies by more than an order of magnitude across tasks: $-12.64$\,pp on $S_3$, $-6.83$\,pp on Parity, and $+4.04$\,pp (not significant) on Dyck-2. We read this as a qualification. The conclusion that ``Mamba induction runs through the short convolution'' is strongest for locally order-sensitive computations in our suite and weaker on weakly local ones, and our long-range bracket-accounting setting does not support it.

%% file: sections/setup.tex
\section{Experimental setup}
\label{sec:setup}

\subsection{Tasks}

\textbf{Parity.} The input is a binary string; the label at position $t$ is the running parity of the prefix. Parity is a commutative, single-bit prefix-update task.

\textbf{$S_3$ permutation composition.} Each input token is one of two generators of the symmetric group $S_3$. The label is the cumulative group product, giving a 6-class task. Unlike Parity, $S_3$ is non-commutative and order-sensitive, but it is still a prefix-update computation: the label at position $t$ depends only on the running group element at $t$.

\textbf{Dyck-$k$.} The input is a bracket sequence with $k$ bracket types, and the label is the current nesting depth, capped at $d_{\max}=10$. The label is the \emph{size} of the stack and not its contents: the prefixes \texttt{([} and \texttt{[(} require different stacks but receive the same label. Depth also satisfies $d_t = d_{t-1} + 1$ if $x_t$ opens and $d_t = d_{t-1} - 1$ if it closes, so it admits a running state in exactly the sense Parity and $S_3$ do, and is a bounded prefix-state update itself. We verified this identity against our data generator with zero error over $17{,}844$ positions. Two consequences follow. First, all three tasks in our suite are prefix updates, so the contrast is extensional: Parity and $S_3$ on one side, Dyck-depth on the other. Second, since $d_t$ equals the number of opening minus closing brackets seen so far, depth is a linear function of the running token counts, which makes it cheap to read from any representation that forms those counts. Appendix~\ref{app:controls} reports a shortcut-controlled follow-up that supervises and probes stack \emph{contents} instead, at positions where the count is uninformative by construction.

\textbf{Bracket counting on code text.} We derive a token-level label from Python functions in CodeSearchNet~\citep{husain2019codesearchnet} by scanning characters and incrementing on \texttt{([\{} and decrementing on \texttt{)]\}}, then labelling each token by the count at its last non-whitespace character. We call this bracket counting rather than nesting depth deliberately. The scan is not a lexer, so brackets inside string literals, comments and regular expressions are counted as if they nested, which they do not. The label distribution is also severely skewed: over $2{,}000$ chunks only $7$ of the $16$ label values occur at all, and increasing the sample does not recover the rest, so the deep classes are genuinely rare rather than undersampled. It is a side observation on non-formal input, and we give balanced accuracy alongside overall accuracy because the two differ by up to $0.55$ here.

All formal tasks train at $L_{\text{train}}=40$ and evaluate at $L \in \{40,60,80,100,120\}$. Training sets contain 50\,000 sequences; test sets contain 5\,000 per length. Code-depth uses fixed-length tokenized code chunks of length 256.

\subsection{Models}

Formal-task experiments use 4-layer models with $d_{\text{model}}=128$. We compare Transformer, Mamba, Mamba-2, LSTM~\citep{hochreiter1997long}, and GRU~\citep{cho2014learning}, plus parameter-matched LSTM/GRU variants and several controls. Transformer, Mamba, and Mamba-2 are approximately parameter-matched at ${\sim}830$K. Full configurations are listed in Appendix~\ref{app:model_config}.

For cross-scale validation we fine-tune pretrained Mamba-130M (24 layers, $d=768$), Pythia-160M~\citep{biderman2023pythia} (12 layers, $d=768$), and Pythia-410M (24 layers, $d=1024$, Dyck only). On Parity and Dyck we replace the embedding and output head with task-specific layers; on code-depth we keep the pretrained embedding and only swap the head. All pretrained experiments use $n=4$ fine-tuning seeds, matching the multi-seed convention in recent mechanistic interpretability work~\citep{arora2025mechanistic}.

\subsection{Measurements}

\textbf{Linear probes.} For each layer we train a logistic regression classifier on frozen hidden states to predict ground-truth state.

\textbf{Zero-ablation.} Given a unit probe direction $\hat{w}$, we intervene by projecting it out of the hidden state, $h' = h - (h \cdot \hat{w})\hat{w}$, and measure the resulting accuracy drop. We compare against random directions and energy-matched controls.

\textbf{Iterative nullspace projection.} The measurement the paper rests on, defined and used in Section~\ref{sec:erasure}. The budget is $100$ directions at $128$ channels and $256$ at pretrained width, and every drop is reported as a function of $k$ as well as at the terminal $k^\ast$.

\textbf{Subspace ablation.} To study within-layer geometry, we remove either (i) top singular directions of the probe weight matrix or (ii) top PCA directions of the hidden states, separating low-rank probe-aligned structure from broader distributed structure. For pretrained Mamba we additionally apply mid-position activation patching (Appendix~\ref{app:methods}).

\textbf{Controls.} We evaluate ALiBi and RoPE positional encodings, a Gated Transformer, and Conv1D-ablated Mamba under a matched pure-PyTorch backend. Appendix sections provide precise implementation details and statistical testing protocol.

%% file: sections/observation.tex
\section{A task-dependent observation}
\label{sec:observation}

Layerwise probe profiles are often read as architectural signatures. They move with the task. On \textbf{Parity} and $S_3$, Mamba and the recurrent baselines stay weak through the early layers and become nearly perfectly readable at the last one, while Transformer accumulates readable state across depth. On \textbf{Dyck} the roles swap: Transformer is at ceiling from layer $0$, and Mamba and the recurrent baselines build the representation progressively (Appendix Fig.~\ref{fig:probes}). $S_3$ is non-commutative and patterns with Parity, so commutativity does not account for the grouping. Length generalization orders the families the same way on all three tasks, recurrent above SSM above Transformer, and capacity and positional-encoding controls do not close the gap (Appendix~\ref{app:extra}).

The contrast is extensional. Dyck-depth is itself a bounded prefix-state update (Section~\ref{sec:setup}), so it groups with Parity and $S_3$ by construction, and what separates the three tasks here is the supervised target. Section~\ref{sec:survive} states the scope this fixes.

Component-level attribution moves with the task in the same way. Removing Mamba's short convolution hurts $S_3$ most, hurts Parity moderately, and has no significant effect on Dyck, with both variants run under a matched pure-PyTorch backend so the comparison is not confounded with kernel implementation (Appendix~\ref{app:conv}). The effect is task-specific even within one architecture family.

%% file: sections/tap.tex
\section{Probing and intervening on one tensor}
\label{sec:tap}

Probing and intervention have to act on the same quantity, and across architectures that is not automatic. We use one tensor throughout: $h_i$, the residual stream immediately after block $i$ and before any final normalization.

The two families expose it differently. A Mamba block returns \texttt{(mixer\_out, residual)}, where \texttt{residual} is the stream accumulated before this block's output is added, so the stream after block $i$ is their sum. Storing \texttt{residual} alone is one block early, and differs from the true stream by a relative $L_2$ of $0.31$ mid-stack and over $0.90$ at either end. On GPT-NeoX, \texttt{output\_hidden\_states} appends the stream before each block and once more after the final layer norm, so its last entry is a normalized tensor and not $h_{n-1}$.

Choices of this kind decide what a patching result means~\citep{zhang2024towards}. Reading one of these while writing the other turns a necessity test into a comparison between two different quantities, and the discrepancy is largest at the ends of the stack, which is where late-layer readability claims are made. Every probe, ablation, subspace ablation and patch in this paper goes through one module that captures $h_i$ with a forward hook and, to intervene, solves for the block output that makes the post-block stream equal the intervened value. We report this because it is the part of our setup most directly reusable by anyone comparing architectures, and because the mismatch is silent: a probe-SVD against PCA contrast at Pythia L11, the one layer where \texttt{output\_hidden\_states} returns the post-final-norm tensor while an ablation writes the block output, compares two tensors and cannot be repaired by restating it.

%% file: sections/notnecessity.tex
\section{Ablating one direction is not a necessity test}
\label{sec:notnec}

Fit a probe at a layer, project its direction out during the forward pass, and read the task accuracy. That measurement is used in both directions~\citep{belinkov2022probing}: a large drop as evidence the computation depends on what the probe read, a drop near zero as evidence that it does not. Both readings rest on one premise, that the ablation removed the target from the layer. This is a default step in probing pipelines rather than the practice of any one paper.

The premise fails. If the target had been removed, a probe fitted from scratch on the ablated activations would lose accuracy. It does not. Refitting at the same layer and under the same protocol recovers the original accuracy in every cell we checked. Dyck Transformer at L0 goes $1.000$ to $1.000$, Parity Mamba at L3 goes $0.987$ to $0.988$. It holds when the entire $K{-}1$ probe row space is removed rather than one axis. The erasure run of Section~\ref{sec:erasure} refits at every step and so repeats the check on its own: after one direction, $58$ of its $60$ cell-seeds are within $0.02$ of where they started, and the two exceptions are both Parity Mamba seeds, only one of them substantially (Fig.~\ref{fig:headline}a).

A probe direction is one linear decoder among many, and two coordinates are enough to see what that costs the intervention. Let a binary target $y$ be written twice, as $h = (y, y)$, and let the fitted probe use the first coordinate. Projecting that coordinate out leaves $h' = (0, y)$: the original probe is dead, and a probe refitted on $h'$ reads $y$ off the second coordinate at no loss. If the downstream computation reads the first coordinate, task accuracy collapses while the refitted probe succeeds; if it reads the second, both survive. Those are the two patterns we observe, on Dyck and on Parity, though the example fixes no claim about the routing the models actually use. In neither case has the ablation erased anything, so in neither case does the drop speak to necessity.

Figure~\ref{fig:headline} shows both cases in the data: on Dyck, deleting the probe's direction costs the model close to a third of its task accuracy while the refitted probe has not moved at all (Fig.~\ref{fig:headline}a against~\ref{fig:headline}c). Two questions are being conflated. A single-direction ablation asks whether the model is fragile to one coordinate; a necessity test asks whether the model can still solve the task once the target itself has been made unavailable. The refit shows that the first intervention never reaches the premise the second question needs. It costs one probe fit, and we ran it for every cell before reporting any ablation number.

Two further artefacts decide whether a single-direction result may be read at all, and both are properties of the instrument rather than of the model. A probe fitted on standardized features is mapped back as $w_h = w_z \oslash \sigma$, which upweights coordinates that barely vary, so the raw-space direction can point along an axis the representation hardly uses. Ablating it then removes little for reasons unrelated to whether the model reads it. Measured against random unit directions at the same layer, most cells clear this check by a wide margin, carrying tens of times the projection energy of a random direction. Pretrained Mamba on Parity averages $1.0$ and never exceeds $2.4$, so its probe direction is no higher-variance than a random one, its $0.02\%$ rank-1 drop was never a causal statement, and the same cell loses almost everything once the subspace is erased. We treat any cell under three times the random-direction energy as unreadable for single-direction ablation and report the subspace result instead. The $3\times$ rule in Table~\ref{tab:pretrained_causal} is a different test, applied to drops.

Alongside plain random controls we also ablate random directions whose projection variance is within $20\%$ of the probe direction's; across all $32$ cells these energy-matched controls stay at or below $0.004$ relative drop, so where a probe direction does damage, the damage is specific to that direction and not to the amount of variance deleted.

The second artefact is the layer index itself. On Dyck every architecture reaches exactly $1.000$ at two to four consecutive layers, so ``the best probe layer'' is an argmax over ties: the layer it lands on changes with the seed, and the measured drop follows the layer rather than the model. Our 4-layer Mamba splits its seeds between L2 and L3, and those two layers give relative drops of $0.12$--$0.14$ and $0.000$. Averaged together they look like cross-seed variance in the model, and are nothing of the kind. Wherever a single layer has to be named we take the shallowest layer whose probe accuracy is within $0.5$ points of the maximum. Parity and $S_3$ have a unique argmax at the final layer for every architecture, so the rule changes nothing there.

%% file: sections/erasure.tex
\section{Subspace erasure as a necessity test}
\label{sec:erasure}

If deleting one direction leaves the target readable, the question is how many directions it takes. Fit a probe, project its direction out, refit on what remains, and repeat until no linear probe beats chance. The directions collected along the way span a subspace whose removal leaves the target linearly undecodable at that layer~\citep{ravfogel2020null}. Ablating that subspace inside the running model is then a necessity test. Table~\ref{tab:inlp} reports it against a random subspace of the same dimension, which is the control that makes the number readable: deleting $k$ arbitrary directions is damaging on its own.

In that table $k^\ast$ is how many directions had to go before no linear probe beat chance, ``probe left'' is what a freshly fitted probe recovers once they are gone, and every drop is relative, (clean $-$ ablated) over (clean $-$ chance). Each cell is read both at its own $k^\ast$ and at a shared $k{=}32$.

\input{tables/tab_inlp}

The result inverts what single-direction ablation reported. On Parity and $S_3$ the rank-1 drops sit between $0.0$ and $4.9\%$ at every architecture and every length (Appendix~\ref{app:rank1ood}). Subspace erasure lands between $0.78$ and $0.98$ relative in nine of the ten cells, against random subspaces of identical dimension costing $0.000$ to $0.039$. Those readable representations are load-bearing, and the zeros were a property of the instrument.

Mamba on Parity is the weak cell at $0.46\pm 0.40$, and one seed inside it is worth isolating. Its Parity subspace erases in $23$ directions, leaves a fresh probe at $0.514$ against chance $0.5$, and costs the model nothing. That is the only place in the grid where a converged erasure meets a zero drop: the target is linearly undecodable at that layer and the model still solves the task.

Dyck must be read at a shared budget rather than at each cell's own $k$. There the erasure runs to the $100$-direction cap in every cell, which is most of a $128$-wide layer, and a random subspace that size is destructive on its own. At $k{=}32$ Transformer and the SSMs still separate cleanly, $0.83$ and $0.56$--$0.67$ against nulls under $0.04$, while LSTM and GRU carry nulls of $0.28$ and $0.14$ and support a weaker claim. Figure~\ref{fig:inlp_curve} gives the full dependence on $k$.

At $130$M and $160$M the same test costs both models most of the task on Parity. Pythia loses $0.71\pm 0.35$ and Mamba $0.89\pm 0.19$, against random subspaces of the same $256$ directions costing $0.024$ and $0.013$ (Table~\ref{tab:inlp_pretrained}). The erasure there is partial rather than complete. $768$ channels hold the target far more redundantly than $128$ do, and only $4$ of $16$ cells reached chance before the budget ran out, so we report the residual probe accuracy alongside the drop. On Dyck the separation is much weaker, with the random control reaching $0.418$ for Pythia, because $256$ directions is a third of the layer.

%% file: tables/tab_inlp.tex
\begin{table*}[ht]
\centering
\footnotesize
\caption{Subspace erasure at the first layer where the target becomes readable, mean $\pm$ std over
$n{=}4$ seeds. Columns are defined in Section~\ref{sec:erasure};
$^{\dagger}$ marks cells that hit the $100$-direction cap before a probe fell to chance.}
\label{tab:inlp}
\setlength{\tabcolsep}{2pt}
\begin{tabular}{@{}llccccccc@{}}
\toprule
& & & & & \multicolumn{2}{c}{\textbf{Drop at $k{=}32$}} & \multicolumn{2}{c}{\textbf{Drop at $k^\ast$}} \\
\cmidrule(lr){6-7}\cmidrule(lr){8-9}
\textbf{Task} & \textbf{Model} & \textbf{Layer} & $k^\ast$ & \textbf{Probe left} & \textbf{probe} & \textbf{random} & \textbf{probe} & \textbf{random} \\
\midrule
Dyck-2 & Transformer & L0 & $100\pm 0$$^{\dagger}$ & $0.314\pm 0.018$ & $0.829\pm 0.112$ & $0.038\pm 0.029$ & $0.932\pm 0.040$ & $0.586\pm 0.043$ \\
 & Mamba & L2 & $100\pm 0$$^{\dagger}$ & $0.446\pm 0.032$ & $0.564\pm 0.198$ & $0.032\pm 0.053$ & $0.699\pm 0.161$ & $0.307\pm 0.060$ \\
 & Mamba-2 & L2 & $100\pm 0$$^{\dagger}$ & $0.402\pm 0.014$ & $0.672\pm 0.078$ & $0.010\pm 0.019$ & $0.850\pm 0.025$ & $0.354\pm 0.061$ \\
 & LSTM & L1--L2 & $100\pm 0$$^{\dagger}$ & $0.422\pm 0.043$ & $0.571\pm 0.132$ & $0.281\pm 0.223$ & $0.857\pm 0.055$ & $0.631\pm 0.144$ \\
 & GRU & L1 & $100\pm 0$$^{\dagger}$ & $0.381\pm 0.048$ & $0.726\pm 0.129$ & $0.140\pm 0.138$ & $0.904\pm 0.053$ & $0.641\pm 0.192$ \\
\midrule
Parity & Transformer & L3 & $96\pm 7$$^{\dagger}$ & $0.656\pm 0.096$ & $0.516\pm 0.419$ & $0.000\pm 0.001$ & $0.945\pm 0.053$ & $0.001\pm 0.001$ \\
 & Mamba & L3 & $72\pm 35$ & $0.512\pm 0.010$ & $0.368\pm 0.459$ & $0.000\pm 0.000$ & $0.464\pm 0.396$ & $0.000\pm 0.000$ \\
 & Mamba-2 & L3 & $66\pm 9$ & $0.517\pm 0.003$ & $0.975\pm 0.045$ & $0.000\pm 0.000$ & $0.972\pm 0.033$ & $0.000\pm 0.000$ \\
 & LSTM & L3 & $100\pm 0$$^{\dagger}$ & $0.859\pm 0.097$ & $0.446\pm 0.228$ & $0.000\pm 0.000$ & $0.965\pm 0.045$ & $0.000\pm 0.000$ \\
 & GRU & L0--L3 & $100\pm 0$$^{\dagger}$ & $0.646\pm 0.086$ & $0.597\pm 0.161$ & $0.000\pm 0.000$ & $0.779\pm 0.404$ & $0.024\pm 0.048$ \\
\midrule
$S_3$ & Transformer & L3 & $96\pm 8$$^{\dagger}$ & $0.194\pm 0.013$ & $0.491\pm 0.114$ & $0.004\pm 0.003$ & $0.968\pm 0.010$ & $0.039\pm 0.024$ \\
 & Mamba & L3 & $97\pm 6$$^{\dagger}$ & $0.222\pm 0.033$ & $0.427\pm 0.165$ & $0.000\pm 0.000$ & $0.974\pm 0.030$ & $0.010\pm 0.004$ \\
 & Mamba-2 & L3 & $77\pm 2$ & $0.184\pm 0.001$ & $0.725\pm 0.089$ & $0.005\pm 0.003$ & $0.981\pm 0.011$ & $0.021\pm 0.008$ \\
 & LSTM & L3 & $100\pm 0$$^{\dagger}$ & $0.561\pm 0.078$ & $0.288\pm 0.095$ & $0.000\pm 0.000$ & $0.925\pm 0.078$ & $0.000\pm 0.000$ \\
 & GRU & L2--L3 & $100\pm 0$$^{\dagger}$ & $0.349\pm 0.080$ & $0.253\pm 0.177$ & $0.000\pm 0.000$ & $0.853\pm 0.093$ & $0.028\pm 0.045$ \\
\bottomrule
\end{tabular}
\end{table*}

%% file: sections/depth.tex
\section{Readability and fragility across depth}
\label{sec:depth}

\begin{figure}[!t]
\centering
\includegraphics[width=\linewidth]{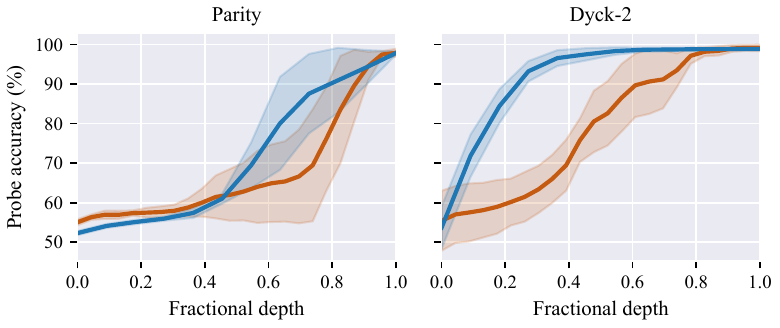}

\vspace{0.3em}
\centerline{\includegraphics[width=0.7\linewidth]{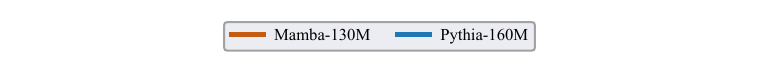}}
\caption{Layerwise probe accuracy on the post-block residual stream of the fine-tuned pretrained models, against fractional depth so the $12$-layer and $24$-layer stacks are comparable. Bands are mean $\pm$ std over $n{=}4$ fine-tuning seeds; chance is $50\%$ on Parity and $9.1\%$ on Dyck-2.}
\label{fig:pretrained_probe_compare}
\end{figure}

A layerwise ablation profile is a common thing to report, and reading one means knowing what its two curves measure. The sweep here uses single-direction ablation, which Section~\ref{sec:notnec} showed is not a necessity test; it measures how much a layer's linear format is disturbed. Mapped across depth, that quantity does not peak where the probe does, which is not what one would guess: the layer that reads a variable best is the natural candidate for the layer whose perturbation costs most.

Fine-tuned Mamba-130M and Pythia-160M reproduce the cross-task probing asymmetry of Section~\ref{sec:observation} (Fig.~\ref{fig:pretrained_probe_compare}): on Parity, Mamba concentrates readable state late while Pythia improves monotonically; on Dyck, Pythia is readable within the first few layers while Mamba builds gradually. Table~\ref{tab:pretrained_causal} sweeps every layer of both models, fitting a probe on $h_i$ and ablating its direction in that same tensor. On Dyck the ablation peak sits strictly below the probe peak in each. Pythia's probe is at ceiling by L11 ($98.8\%$) but the largest drop is at L7, $30.2\pm 19.0$ points against a random-direction null of $0.4$; Mamba's probe peaks at L23 ($99.0\%$) where ablation costs $0.67\pm 0.64$, and its own peak is at L19 with $46.3\pm 19.9$. Divided by depth the two peaks land at $0.64$ and $0.83$, one phenomenon at two depths rather than two architecture-specific failure modes. Pythia-410M keeps the shape at $2.5\times$ the width: probe peak at L22 ($98.8\%$), ablation peak at L9, normalized depth $0.39$. Its peak drop is $21.2\pm 23.0$ against $30.2\pm 19.0$ at $160$M, so we read no scale trend in the magnitude, only in the arrangement.

Single-direction ablation gives a null on Parity and $S_3$ at this scale. Pythia's largest Parity drop is $3.6\pm 6.4$ at L11, a standard deviation larger than the mean; Mamba has no layer on either task whose drop clears three times the drop from a random direction there, and its L23 Parity direction carries no more variance than a random one, so that zero says nothing either way (Section~\ref{sec:notnec}). Sufficiency behaves differently from necessity. Mid-position patching~\citep{vig2020investigating,meng2022locating} at Mamba L22 recovers $97.5\pm 4.4\%$ of the clean--corrupted logit gap on Parity and $98.4\pm 0.7\%$ on Dyck (Table~\ref{tab:patching}). The stream at that site nearly fixes the readout even where deleting any single direction in it costs nothing.

Erasure asks whether the model can work without a set of linearly decodable directions; it does not locate where the computation happens. A per-layer erasure sweep would at least let that necessity signal be compared across depth, and our budget does not reach chance even at one layer. The sweep reported here answers the narrower question of single-direction fragility, and it gives the comparison the practical point needs: the layer with the highest probe accuracy is not the layer with the largest ablation response, in every pretrained model we ran, and a profile is fixed by neither measurement alone. \citet{hase2023localization} reach the same shape from a different pair of measurements, finding that where causal tracing localizes a fact does not say which layer it is best to edit.

Bracket counting on real Python, and a frozen-backbone check at $1.4$B, both reproduce the Dyck-side ordering (Appendices~\ref{app:pretrained} and~\ref{app:controls}). The four-layer dissociation, top-of-stack content being readable from content-supervised models but not from size-supervised ones, does not survive there once the comparison is made at matched task accuracy: probed at the first checkpoint where each model reaches a fixed accuracy target, the difference is gone.

%% file: sections/survive.tex
\section{Limitations}
\label{sec:survive}
\label{sec:limitations}

The task grouping of Section~\ref{sec:observation} holds for these three tasks with these supervised targets, and it is extensional by construction: Dyck-depth equals opening minus closing brackets, which we verified against the data generator with zero error over $17{,}844$ positions, so it is a bounded prefix-state update exactly like Parity and $S_3$. Which property of a task produces the grouping is open, and Appendix~\ref{app:controls} reports the tests that fix this scope, each against the rule it was judged by~\citep{hofman2023pre}.

Erasure separates from a matched random subspace in every cell we report, and two things bound how far that reading goes. At pretrained scale the erasure is partial, since only $4$ of $16$ cells reached chance inside the budget, so the drops there are a lower bound on what a complete erasure would cost rather than an estimate of it. And cross-seed spread is wide throughout, $0.46\pm 0.40$ at 4-layer Mamba on Parity and $\pm 0.35$ at pretrained Pythia, which is why we read the effect at the level of the grid and not of a single cell.

What the erasure removes is linear decodability, which is what a probe reads and therefore what the claim is about. A quantity can still survive nonlinearly: \citet{ravfogel2022kernel} kernelize the erasure objective, block one nonlinear adversary, and find that the protection does not transfer to another, so exhaustive removal remains open. Deleting a subspace likewise shows that a model cannot solve the task without the directions it removed, which is the claim we make, and not that those directions are where the computation lives, since a quantity can be readable in one basis and used in another that overlaps it.

Pythia-410M is our largest sweep, bounded by the cost of a full fine-tune at every layer; the algebraic axis we vary is commutativity, and associativity and identity-element variants, along with hybrids such as Jamba~\citep{lenz2025jamba}, are open.

%% file: sections/conclusion.tex
\section{Conclusion}
\label{sec:conclusion}

Ablating a probe direction does not remove what the probe read, so the number it produces cannot support a claim about necessity in either direction. Erasing the subspace does remove it from every linear probe we can fit, and under that measurement the readable representations in our models are load-bearing. The layer that reads best is also not the layer where deleting the probe direction costs most, in every pretrained model we ran. Three things follow for practice. Report a layerwise profile together with the target it was measured under, since profiles measured under different targets are not comparable (Appendix~\ref{app:controls}). Refit the probe on the ablated activations before reading an ablation number, which costs one probe fit. And check that probe and intervention act on the same tensor, which across architectures is not automatic. Each of the three costs less than the measurement it corrects.

%% file: sections/appendix.tex
\section{Models, tasks, and training details}
\label{app:model_config}

\input{tables/tab1_config}

Table~\ref{tab:config} lists all formal-task model variants. All formal experiments use $4$ layers and $d_\text{model}=128$; Transformer, Mamba (S6), and Mamba-2 (SSD) are approximately parameter-matched at ${\sim}830$K. Optimization uses AdamW with a cosine schedule and task-specific learning rates (Parity: lr~$3\times10^{-4}$, batch~64, 30~epochs; Dyck and $S_3$: lr~$4\times10^{-4}$, batch~128, 30~epochs with 3-epoch warmup).

\paragraph{Recurrent baselines.} LSTM~\citep{hochreiter1997long} and GRU~\citep{cho2014learning} use four stacked pre-norm blocks with residual connections and $d_\text{model}=128$. Parameter-matched variants are LSTM-m ($d=160$, ${\sim}827$K) and GRU-m ($d=185$, ${\sim}828$K). In-distribution accuracy exceeds $99\%$ for all recurrent variants and seeds.

\paragraph{Gated Transformer.} We insert a GRU-style gate after each attention sub-layer: $g = \sigma(W_g [x; \mathrm{attn}(x)] + b_g)$ and output $g \odot \mathrm{attn}(x) + (1{-}g)\odot x$. The gate is position-wise. It is $\mathrm{Linear}(2d_\text{model}, d_\text{model})$, so it adds $32{,}896$ parameters per layer. Total parameters are held roughly constant by reducing $d_\text{ff}$ from $512$ to $384$, which means this variant changes the feed-forward width as well as adding the gate. It therefore does \emph{not} isolate within-step gating, and we read it only as evidence that this particular gated variant does not close the Transformer's gap, not as an attribution to gating.

\paragraph{Mamba-2.} We use the SSD backbone with $d_\text{state}=24$, $\text{expand}=4$, $\text{headdim}=64$, $\text{ngroups}=1$, totaling $830{,}242$ parameters. The $D_\text{total} = 2d_\text{inner} + 2\cdot\text{ngroups}\cdot d_\text{state} + n_\text{heads}$ divisibility-by-8 constraint is satisfied with $d_\text{inner}=512$, $d_\text{state}=24$, $n_\text{heads}=8$.

\paragraph{Pretrained models.} Mamba-130M (24 layers, $d=768$, ${\sim}90.5$M non-embedding parameters) and Pythia-160M (12 layers, $d=768$, ${\sim}85$M non-embedding parameters) are loaded from HuggingFace checkpoints. Fine-tuning uses AdamW at lr~$10^{-5}$, weight decay~$0.01$, cosine schedule, batch~$64$, 15--20 epochs, with early stopping at $90$--$95\%$ in-distribution accuracy ($95\%$ on Parity/Dyck/$S_3$, $90\%$ on code-depth). We replace the embedding and output head with task-specific layers on Parity/Dyck, and retain the pretrained embedding while swapping only the prediction head on code-depth. All pretrained experiments use $n=4$ seeds.

\paragraph{Tasks and data generation.}
\label{app:tasks}

\textbf{Parity.} Binary inputs $x_t \in \{0,1\}$; label $y_t = \bigoplus_{i\le t} x_i$. A commutative prefix-update task with a single-bit latent state. \textbf{$S_3$ permutation composition.} The symmetric group $S_3$ has $|S_3|=6$ elements. We use two generators $\sigma_0=(1\,2)$ (transposition) and $\sigma_1=(1\,2\,3)$ (cyclic rotation). Each input position is sampled uniformly from $\{\sigma_0, \sigma_1\}$. The per-position label is the cumulative group product $g_t = g_{t-1}\circ x_t$ starting from the identity, giving a 6-class non-commutative prefix-update task. Non-commutativity is verified by $\sigma_0\circ\sigma_1\neq\sigma_1\circ\sigma_0$. The $6\times 6$ multiplication table is precomputed. \textbf{Dyck-$k$.} Random walks on a bracket stack with push probability $0.55$ and pop probability $0.45$, subject to $d_\text{max}=10$ and a closure constraint that reserves enough positions to close all open brackets. Bracket types are uniform over $\{1,\ldots,k\}$; labels are per-position stack depths ($0$ to $d_\text{max}$, 11 classes). The main text uses Dyck-2; Dyck-3 and Dyck-4 are tested in Appendix~\ref{app:extra}. All three formal tasks use $50{,}000$ training and $5{,}000$ test sequences per length, at $L_\text{train}=40$ and evaluation lengths $L \in \{40,60,80,100,120\}$. \textbf{Code-depth.} A token-level nesting-depth task derived from Python functions in CodeSearchNet~\citep{husain2019codesearchnet}. Each function is tokenized with the pretrained model's tokenizer, chunked to length~256, and labeled with the bracket nesting depth at each token's last non-whitespace character (11 classes), yielding a real-code analogue of the Dyck-depth task.

\section{Probing, intervention, and statistical testing}
\label{app:methods}

\paragraph{Linear probes.} One multinomial logistic regression per layer, shared across all supervised positions rather than fitted per position, so that a direction found by the probe is the direction an ablation at that layer removes. Sequences are split $80/20$ before positions are flattened, so no position of a test sequence appears in training. Features are standardized with statistics from the training split only. We use $\ell_2$ regularization at $C=1.0$ and allow up to $10{,}000$ iterations, and we record \texttt{n\_iter\_} for every fit; a cell in which any layer fails to converge is marked failed rather than reported. The regularization strength was chosen by measurement on the hardest cell, $n_\text{train}=64{,}000$ over $11$ classes: $C=0.1$ cost accuracy without fixing convergence, and $C=1/n$ over-regularized badly, taking one layer from $0.978$ to $0.418$. Accuracy is reported overall, balanced, and per position. A probe direction learned in standardized space is mapped back as $w_h = w_z \oslash \sigma$ before it is used for anything, since $w_z\!\cdot\!z$ equals $(w_z \oslash \sigma)\!\cdot\!h$ up to a constant and ablating $w_z$ itself removes a different direction than the one that was scored.

A two-layer MLP probe (hidden~256, ReLU, dropout~0.1) trained on the same states gives negligible improvement at key layers (Table~\ref{tab:mlp}), suggesting that the linear view is not hiding substantial nonlinear structure at the layers we report on.

\begin{table*}[ht]
\centering
\footnotesize
\caption{Linear vs.\ two-layer MLP probe accuracy at the key layer of each (task, architecture) pair (mean $\pm$ std across $n{=}4$ seeds). The MLP probe closes at most $0.75 \pm 0.22$\,pp of the linear-probe gap, suggesting that the layer we report as ``where state becomes readable'' is not hiding a substantial nonlinear component.}
\label{tab:mlp}
\setlength{\tabcolsep}{6pt}
\begin{tabular}{@{}llccc@{}}
\toprule
\textbf{Task} & \textbf{Model} & \textbf{Layer} & \textbf{Linear} & \textbf{MLP} \\
\midrule
Parity & Mamba       & L3 & $99.79 \pm 0.17\%$ & $99.96 \pm 0.02\%$ \\
Parity & Transformer & L3 & $99.93 \pm 0.14\%$ & $99.95 \pm 0.11\%$ \\
Dyck-2 & Mamba       & L1 & $99.94 \pm 0.04\%$ & $100.00 \pm 0.00\%$ \\
Dyck-2 & Transformer & L0 & $99.21 \pm 0.21\%$ & $99.96 \pm 0.03\%$ \\
\bottomrule
\end{tabular}
\end{table*}

The maximum key-layer gap is $0.75 \pm 0.22$\,pp (Dyck Transformer L0, $n{=}4$); non-key (early) layers show larger gaps (e.g.\ Dyck Mamba L0: $6.05 \pm 2.11$\,pp at $n{=}4$), pointing to partially nonlinear encodings that become linearized in deeper layers.

\paragraph{Zero-ablation.} Given a unit probe direction $\hat w$, we intervene by projecting it out of the hidden state at the target layer, $h' = h - (h\cdot\hat w)\hat w$, and measure the accuracy drop. For Transformer blocks the hook intercepts the residual stream after both attention and FFN; for Mamba blocks it intercepts after the block residual addition.

\paragraph{Energy-matched controls.} To address the possibility that ``any high-variance direction hurts,'' we compare against random unit vectors whose projection energy $\text{Var}(H\hat w)$ matches the probe within $\pm 20\%$. We draw $10{,}000$ candidates and select the $200$ closest in energy (widening the band if fewer than $200$ match). Across $n{=}4$ seeds per cell, mean energy-matched drops remain small: Parity Mamba $0.0003 \pm 0.0002\%$ (probe $19.16 \pm 11.66\%$), Parity Transformer $-0.0002 \pm 0.0002\%$ (probe $32.33 \pm 2.94\%$), Dyck Mamba $0.0034 \pm 0.0061\%$ (probe $69.04 \pm 2.20\%$), Dyck Transformer $0.0001 \pm 0.0001\%$ (probe $64.45 \pm 3.39\%$). The probe direction exceeds all $200$ controls in every case ($p \le 0.005$).

\paragraph{OOD causal ablation protocol.} (1)~Train a logistic regression probe on frozen hidden states at $L=40$ to select the best layer and extract $\hat w$. (2)~For each $L\in\{40,80,100,120\}$, generate a fresh $3{,}000$-sample test set (seed offset by length). (3)~Apply the ablation $h'=h-(h\cdot\hat w)\hat w$ and measure accuracy drop. (4)~Compare against $500$ random unit directions to obtain an empirical permutation $p$-value.

\paragraph{Subspace ablation.} For a finer view of within-layer geometry we remove either the top-$k$ right-singular directions of the probe weight matrix or the top-$k$ PCA directions of the hidden states. The pretrained picture is not uniform across (model, task) cells, and we report all three cases for which we have data:

\emph{Formal Parity and Dyck (4-layer formal-task models).} On formal Parity, rank-$1$ probe-SVD ablation captures the full drop for both Mamba and Transformer; on formal Dyck, early Mamba layers require somewhat larger $k$. The probe-aligned direction is behaviourally consequential there, though its removal does not erase the target.

A probe-SVD against PCA contrast is not available for Pythia code-depth at L11, and the reason is worth stating. L11 is Pythia's final block, and it is the one layer where \texttt{output\_hidden\_states} returns the tensor after the final layer norm while an ablation writes the block output, so the probe and the intervention were reading and writing different tensors at exactly that site. Code-depth enters this paper through layerwise probing only (Appendix~\ref{app:pretrained}).

Low-rank ablation at pretrained scale is reported in Tables~\ref{tab:pretrained_causal} and~\ref{tab:rowspace_specificity} and by erasure in Table~\ref{tab:inlp_pretrained}, all on the unified tap. We do not report top-$k$ probe-SVD or PCA drops for the pretrained models: on those models the quantity survives the deletion (Section~\ref{sec:notnec}), so a small drop at a chosen $k$ is a statement about the rank of the cut rather than about the representation.

\paragraph{Mid-position activation patching ($n{=}4$ seeds).} As a complementary causal test we run mid-position activation patching: capture the layer-$L$ output at the readout position $t_\text{mid}$ from a clean-input forward, replace the same $(L, t_\text{mid})$ slice in a corrupted-input forward via a forward hook, and measure logit recovery $r=(\ell_\text{patched}-\ell_\text{corrupt})/(\ell_\text{clean}-\ell_\text{corrupt})$ where $\ell$ is the logit at the clean label. Pairs are constructed by matching test sequences with different ground-truth labels at $t_\text{mid}$ (200 pairs per seed). Table~\ref{tab:patching} reports cross-seed means. Patching at L22 (with one residual block downstream of the patched site) recovers $97.5\pm 4.4\%$ of the clean--corrupt logit gap on Parity and $98.4\pm 0.7\%$ on Dyck-2; the patched accuracy at the clean label tracks the unperturbed clean accuracy. Patching at L23 trivially gives $1.000\pm 0.000$ recovery (the final block's output feeds only the read-only norm and head, so a single-position substitution there is mathematically equivalent to running clean from L23 onwards) and serves as a sanity check on the protocol. The L22 cell is the informative one: subsequent SSM/conv layers see corrupted-history states at all other positions but a clean residual at the readout, and recovery is still near complete.

\begin{table*}[ht]
\centering
\footnotesize
\caption{Mamba-130M mid-position activation patching at the readable final layers (mean $\pm$ std across $n{=}4$ fine-tuning seeds, $200$ paired samples per seed). Recovery $r=(\ell_\text{patched}-\ell_\text{corrupt})/(\ell_\text{clean}-\ell_\text{corrupt})$ is evaluated at the clean label. L23 is a sanity check (trivially $1.000$ because no further computation follows in the residual stream); L22 is the informative cell, where one downstream block still processes corrupted history at all non-readout positions.}
\label{tab:patching}
\setlength{\tabcolsep}{6pt}
\begin{tabular}{@{}llccc@{}}
\toprule
\textbf{Task} & \textbf{Layer} & \textbf{Recovery $r$} & \textbf{Patched acc @ $y_\text{clean}$} & \textbf{Clean acc} \\
\midrule
Parity & L22 & $0.975\pm 0.044$ & $96.62\pm 5.79\%$ & $99.37\%$ \\
Parity & L23 & $1.000\pm 0.000$ & $99.37\pm 0.75\%$ & $99.37\%$ \\
Dyck-2 & L22 & $0.984\pm 0.007$ & $96.88\pm 2.32\%$ & $97.50\%$ \\
Dyck-2 & L23 & $1.000\pm 0.000$ & $97.50\pm 1.78\%$ & $97.50\%$ \\
\bottomrule
\end{tabular}
\end{table*}

The two intervention families together clarify the Mamba pretrained representation. Substituting the clean L22 hidden state at the readout position nearly restores the clean prediction even with the corrupted history left untouched, so that state is close to sufficient for this readout intervention. No low-rank cut at that layer is individually fragile, though: neither a probe-aligned axis (rank-$1$) nor a wide PCA subspace (rank-$\le 32$). Neither cut is an erasure, so by Section~\ref{sec:notnec} that says nothing either way about necessity. The probe direction is a poor proxy for the mechanism here.

\subsection{Rank-1 ablation out of distribution}
\label{app:rank1ood}

\begin{figure}[!t]
\centering
\begin{subfigure}[t]{0.32\linewidth}
    \centering
    \includegraphics[width=\linewidth]{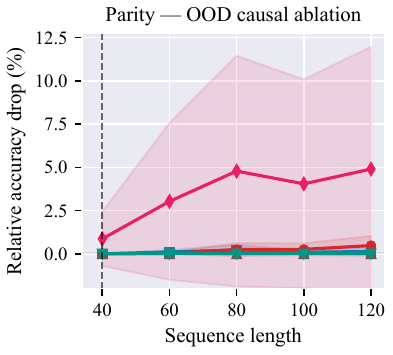}
    \caption{Parity}
    \label{fig:ood_parity}
\end{subfigure}
\hfill
\begin{subfigure}[t]{0.32\linewidth}
    \centering
    \includegraphics[width=\linewidth]{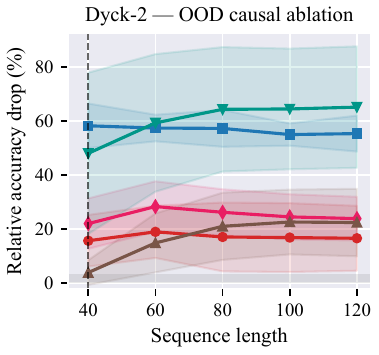}
    \caption{Dyck-2}
    \label{fig:ood_dyck}
\end{subfigure}
\hfill
\begin{subfigure}[t]{0.32\linewidth}
    \centering
    \includegraphics[width=\linewidth]{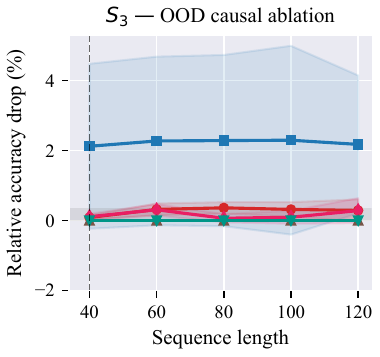}
    \caption{$S_3$}
    \label{fig:ood_s3}
\end{subfigure}

\vspace{0.3em}
\centerline{\includegraphics[width=0.75\linewidth]{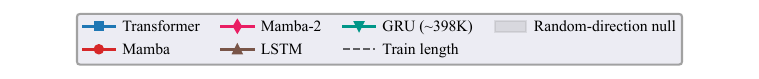}}
\caption{Rank-1 OOD ablation: the probe direction is fitted at the training length $L{=}40$ and then projected out at unseen lengths, $n{=}4$ seeds per cell. The grey band is the largest random-direction drop over all cells of the panel. Drops are relative to (clean $-$ chance) at each length, so they are not inflated by the models' own length degradation.}
\label{fig:ood_ablation}
\end{figure}

\input{tables/tab_inlp_pretrained}

Table~\ref{tab:inlp_pretrained} repeats the erasure at $130$M and $160$M. The measurement still separates on Parity, where erasure costs $0.71\pm 0.35$ for Pythia and $0.89\pm 0.19$ for Mamba against random subspaces of the same $256$ directions costing $0.024$ and $0.013$. On Dyck it separates much less: the random control reaches $0.418$ for Pythia, because $256$ directions is a third of a $768$-channel layer and deleting that much arbitrary structure is damaging by itself, the same effect that governs the $k^\ast$ columns of Table~\ref{tab:inlp}.

Two cautions belong with this table. The budget does not suffice at width: $768$ channels hold the target far more redundantly than $128$ do, and only $4$ of $16$ cells reached chance before the cap, so the erasure is partial and the residual probe accuracy says how partial. And the cross-seed spread is large, $\pm 0.35$ on Pythia Parity, which is the same order as the $0.46\pm 0.40$ that made the 4-layer Mamba cell look exceptional. On four seeds that cell is not obviously outside the range these measurements produce anywhere, and we do not read it as an architecture-specific effect.

\begin{figure}[!tbp]
\centering
\includegraphics[width=\linewidth]{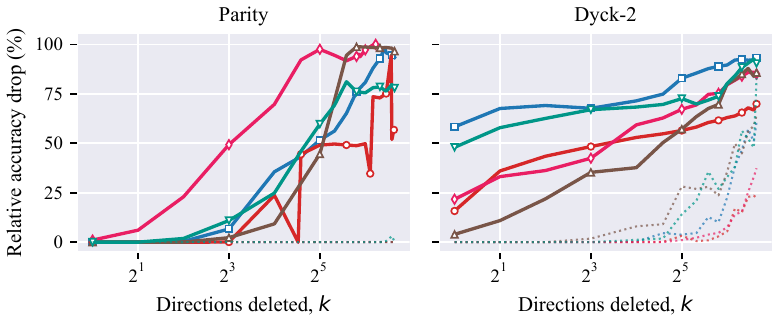}

\vspace{0.3em}
\centerline{\includegraphics[width=0.95\linewidth]{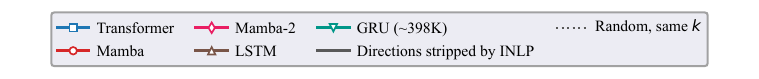}}
\caption{Relative task drop against the number of deleted directions, mean over seeds, at each cell's first readable layer. Solid lines delete the directions iterative nullspace projection strips, in that order; dotted lines delete the same number of random directions at the same layer.}
\label{fig:inlp_curve}
\end{figure}

\subsection{Statistical testing} Two different tests appear in the paper and they answer different questions, so we name which is used where. Ablation results compare one direction or subspace against random ones drawn at the same layer, which is a within-model comparison against an empirical null, and they use one-sided permutation tests with the Phipson--Smyth correction~\citep{phipson2010permutation} $p = (n_\geq + 1)/(N+1)$; with $N=500$ random baselines the smallest attainable value is $p\approx 0.002$. Family-wise error across the 8 primary ablation tests (2 tasks~$\times$~2 models~$\times$~2 control types) is controlled by the Holm procedure~\citep{holm1979simple}, and all corrected values are $p\le 0.02$. The Conv1D ablation instead compares two sets of independently trained models, so it uses Welch's $t$-test on unequal group sizes: $8$ Mamba (Pure) seeds against $4$ No-Conv seeds on Dyck-2, and $4$ against $4$ on Parity and $S_3$. Dispersion is reported as the sample standard deviation with $\mathrm{ddof}{=}1$ throughout, including in the generated tables. Length generalization uses mean~$\pm$~std over $n=4$ seeds; per-layer visualizations show a single seed for clarity, with $p$-values computed across all seeds.

\section{Component-level analyses: Conv1D and attention heads}
\label{app:conv}

\paragraph{Conv1D ablation.}
No-Conv Mamba sets $d_\text{conv}=1$, replacing the depthwise 1D convolution (kernel size~4) with a pointwise convolution (kernel size~1). The pointwise weight remains learnable. Full Mamba uses $384\times 4 + 384 = 1{,}920$ Conv1D parameters per layer; No-Conv uses $384\times 1 + 384 = 768$, a reduction of $4{,}608$ parameters total, which is $0.52\%$ of the $879$K-parameter pure-backend model. Both Full and No-Conv variants use a matched pure-PyTorch parallel scan instead of the official CUDA kernel, because the fused kernel requires $d_\text{conv}\ge 2$. Under the pure backend, Mamba (Pure) reaches $75.37\pm 3.51\%$ on Parity ($n=4$) and $51.43\pm 2.10\%$ on Dyck ($n=8$), close to the official-kernel numbers ($77.27\%$ and $52.38\%$). The kernel choice therefore contributes only ${\sim}2$\,pp, and the Conv1D differences we report are not confounded with kernel implementation.

\paragraph{Training budget for ablation controls.}
On Parity, the Conv1D ablations (\texttt{mamba\_noconv}, \texttt{mamba\_pure}) and the Transformer positional-encoding ablations (\texttt{transformer\_alibi}, \texttt{transformer\_rope}) are trained with a longer schedule than the vanilla Parity defaults: $40$ epochs (vs $30$), batch size $128$ (vs $64$), peak learning rate $4\times 10^{-4}$ (vs $3\times 10^{-4}$), and a $5$-epoch warmup (vs none). These ablation variants are intentionally given the larger budget so as not to disadvantage them. Even with the extra training budget they reproduce the same architecture-task ranking as the vanilla setup, so the additional compute does not change the qualitative conclusion: kernel choice, Conv1D, and positional encoding are not the explanation for the Parity-side architecture asymmetry. All Dyck and $S_3$ vanilla configurations and the corresponding ablations on these two tasks share a single training schedule per task (Dyck: $30$ epochs, batch $128$, lr $4\times 10^{-4}$, $3$-epoch warmup; $S_3$: $30$ epochs, batch $64$, lr $3\times 10^{-4}$, no warmup), so no analogous note is needed there.

\paragraph{Attention head analysis on Dyck.}
\label{app:attn_heads}
We extract per-head attention weights from Transformer Layer~0 on Dyck and measure how much of each head's attention lands on unmatched open brackets up to each query position.

\begin{table}[ht]
\centering
\footnotesize
\caption{Per-head open-bracket attention score on Dyck-2 Layer $0$, averaged across $n=4$ seeds ($500$ samples each). ``Spread'' is the mean within-seed head range ($\max-\min$ across heads, averaged over seeds); the per-head scores shown are the across-seed means. No single head specializes in bracket counting under learned or gated positional encodings; ALiBi and RoPE induce more head variability but still no dedicated bracket head.}
\label{tab:attn_heads}
\setlength{\tabcolsep}{6pt}
\begin{tabular}{@{}lccccc@{}}
\toprule
\textbf{Variant} & \textbf{H0} & \textbf{H1} & \textbf{H2} & \textbf{H3} & \textbf{Spread} \\
\midrule
Vanilla & $0.222$ & $0.221$ & $0.222$ & $0.222$ & $0.003$ \\
Gated   & $0.223$ & $0.223$ & $0.224$ & $0.223$ & $0.004$ \\
ALiBi   & $0.217$ & $0.185$ & $0.213$ & $0.223$ & $0.043$ \\
RoPE    & $0.198$ & $0.195$ & $0.199$ & $0.202$ & $0.038$ \\
\bottomrule
\end{tabular}
\end{table}

No head specializes in bracket counting. In the vanilla Transformer, all four heads produce near-identical uniform prefix aggregation with a recency bias (spread $<0.005$), and the FFN then extracts the depth count. Head-level redundancy and direction-level fragility coexist here, which is worth stating because the two are easy to conflate. Every head writes into the same Layer~0 depth code, so no head is individually necessary; the code itself is, and deleting the subspace that carries it costs $87.8$ points of relative accuracy against $1.0$ for a random subspace of the same size (Table~\ref{tab:inlp}). Redundancy across components does not imply redundancy across directions.

\section{Pretrained experiments: layer sweeps, causal tests, and code-depth}
\label{app:pretrained}

We sweep every layer rather than naming a best probe layer, because that choice is an argmax, and an argmax over a saturated profile is decided by ties. On Dyck this is not a corner case: every architecture we train reaches exactly $1.000$ at two to four consecutive layers, and at pretrained scale Pythia is at or near ceiling from L6 onward. The layer the argmax lands on then changes with the seed, and the measured drop follows the layer rather than the model. Our 4-layer Mamba picks L2 under seeds $42$ and $300$ and L3 under seeds $100$ and $200$, giving relative drops of $0.12$--$0.14$ at L2 against $0.000$ at L3; reported as a mean it looks like large cross-seed variance in the model, and it is nothing of the kind. Wherever a single layer has to be named we take the shallowest layer whose probe accuracy is within $0.5$ points of the maximum, which is deterministic and is also the layer the surrounding claims are about, namely where the quantity first becomes linearly readable. Parity and $S_3$ have a unique argmax at the final layer for every architecture, so the rule changes nothing there. Everything else is reported as a full sweep.

\paragraph{Layerwise ablation sweep.} For every fine-tuned model-task pair, we apply zero-ablation at every layer using that layer's own probe direction, on the same tensor the probe was fitted on, and compare against random unit directions ablated at the same layer. Table~\ref{tab:pretrained_causal} gives the probe peak and the ablation peak side by side.

\input{tables/tab_pretrained_causal}

Both models put the ablation peak strictly before the probe peak on Dyck, at L7 of $11$ for Pythia and L19 of $23$ for Mamba. Mamba's peak is the larger of the two, $46.29\pm 19.85$ against $30.24\pm 19.00$, so the earlier reading in which pretrained Mamba simply has no fragile layer was a consequence of ablating a tensor the probe had not been fitted on. What Mamba does have is a sharp collapse over its last four blocks, from $46.29$ at L19 to $0.67$ at L23, while probe accuracy over the same span climbs to $99.0\%$.

Ablating the probe's full $K{-}1$ row space needs a random subspace of the same dimension to be read at all, and Table~\ref{tab:rowspace_specificity} gives that comparison. On Parity and $S_3$ the probe's own directions are the \emph{less} damaging choice at almost every layer, $22$ of $24$ for Mamba Parity and $10$ of $12$ for Pythia Parity, and only in the last two or three blocks does the probe subspace pull ahead of a random one. On Dyck the crossover comes much earlier, at L1 for Pythia and L13 for Mamba. A subspace-ablation number quoted without its matched null is therefore not interpretable in the shallow half of either network.

\input{tables/tab_rowspace_specificity}

\begin{figure}[!tbp]
\centering
\includegraphics[width=0.95\linewidth]{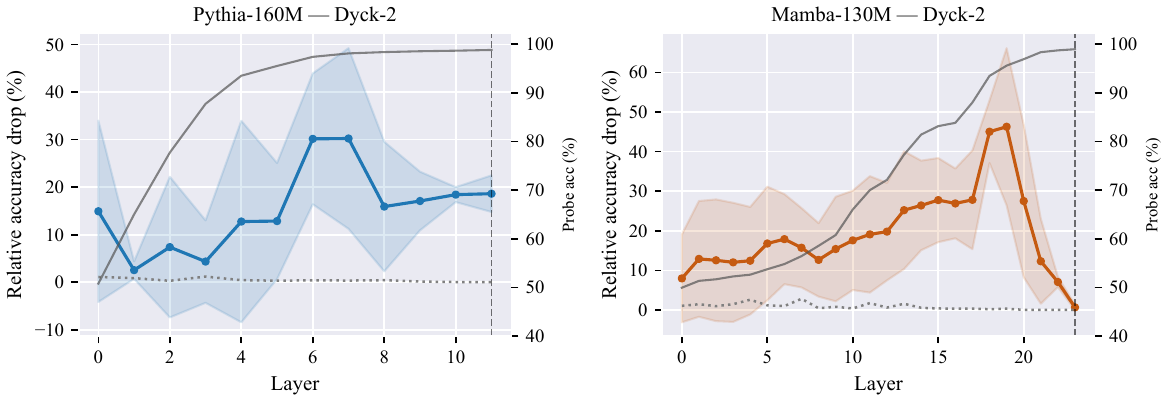}

\vspace{0.3em}
\centerline{\includegraphics[width=0.95\linewidth]{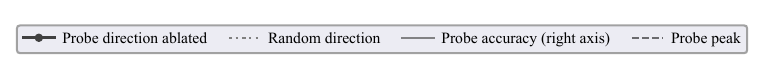}}
\caption{Layerwise single-direction ablation on Dyck-2, mean over four seeds, with the matched random-direction null and probe accuracy on the right axis. Both models place their ablation peak below the probe peak, and both fall back to near zero at the last block, where the probe is at its most accurate.}
\label{fig:pretrained_causal_summary}
\end{figure}

\begin{figure}[!tbp]
\centering
\includegraphics[width=0.62\linewidth]{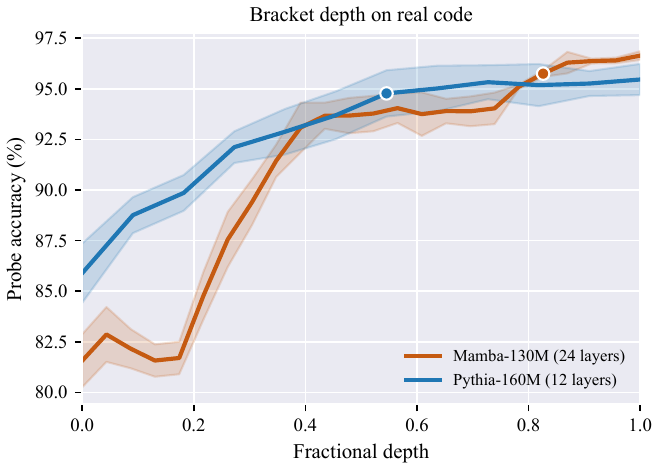}
\caption{Bracket-depth probing on real Python, mean $\pm$ std over $n{=}4$ fine-tuning seeds, against fractional depth so the $12$-layer and $24$-layer stacks are comparable. The marked point on each curve is where that model first comes within one accuracy point of its own ceiling.}
\label{fig:code_depth_probe}
\end{figure}

The code-depth task carries the picture from formal languages to real Python, tokenizing each function and labeling tokens with the bracket nesting depth at the last non-whitespace character. It is not a natural-language benchmark, and its labels are skewed toward shallow nesting, which is why both models already read depth well above chance at the input layer: $85.9 \pm 1.5\%$ for Pythia-160M and $81.6 \pm 1.3\%$ for Mamba-130M over $n{=}4$ fine-tuning seeds. Endpoint accuracies are therefore the wrong summary, and the depth at which each model gets there is the right one. Pythia comes within one point of its own ceiling at $55\%$ of its stack, Mamba at $83\%$ of its (Fig.~\ref{fig:code_depth_probe}), which is the same early-versus-gradual ordering the formal Dyck profiles show. What the task adds is that the ordering is not an artifact of procedurally generated sequences; what it cannot support, given the label skew, is any claim about absolute readability.

\begin{figure}[!t]
\centering
\begin{subfigure}[t]{0.32\linewidth}
    \centering
    \includegraphics[width=\linewidth]{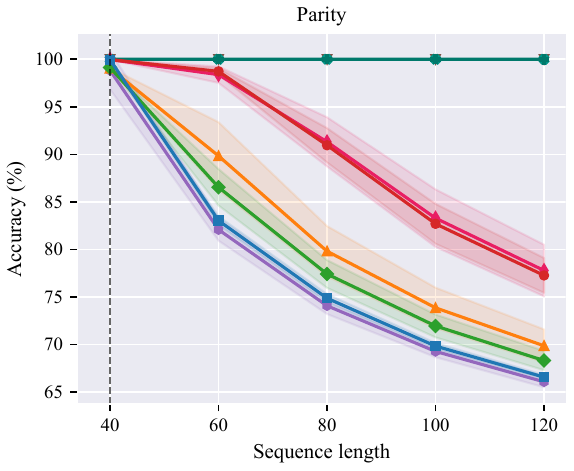}
    \caption{Parity}
    \label{fig:lengthgen_parity}
\end{subfigure}
\hfill
\begin{subfigure}[t]{0.32\linewidth}
    \centering
    \includegraphics[width=\linewidth]{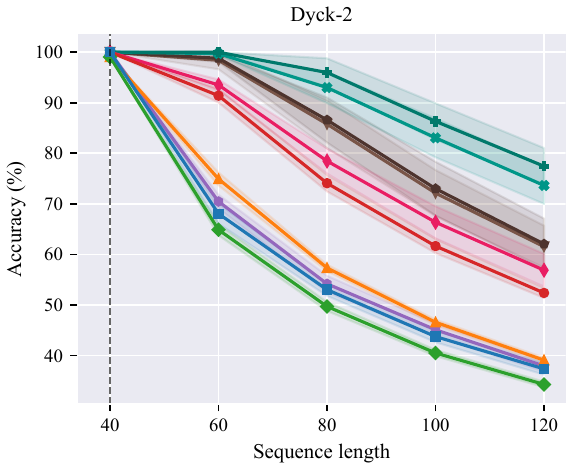}
    \caption{Dyck-2}
    \label{fig:lengthgen_dyck}
\end{subfigure}
\hfill
\begin{subfigure}[t]{0.32\linewidth}
    \centering
    \includegraphics[width=\linewidth]{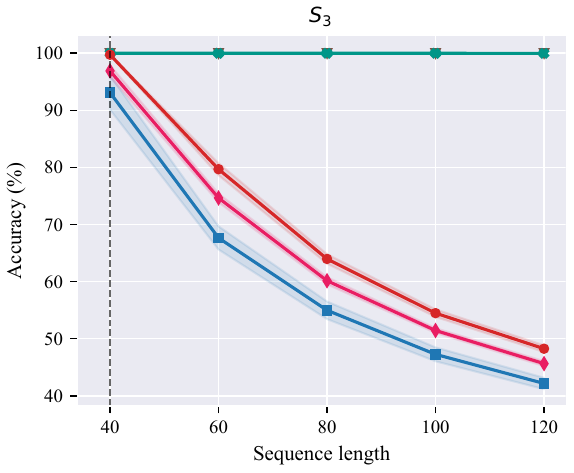}
    \caption{$S_3$}
    \label{fig:lengthgen_s3}
\end{subfigure}

\vspace{0.3em}
\centerline{\includegraphics[width=0.95\linewidth]{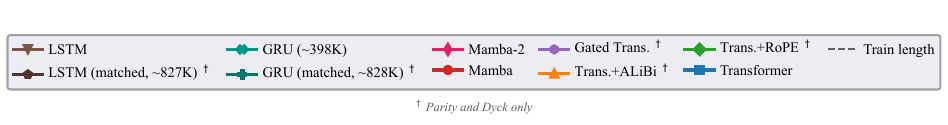}}
\caption{Length generalization at unseen lengths (mean $\pm$ std, $n=4$ seeds). LSTM and GRU curves overlap at the $100\%$ ceiling on Parity and $S_3$. Legend entries marked $^\dagger$ (matched LSTM/GRU, ALiBi, RoPE, Gated Transformer) are confound controls, run only on Parity and Dyck.}
\label{fig:lengthgen}
\end{figure}

\begin{figure}[!t]
\centering
\begin{subfigure}[t]{0.32\linewidth}
    \centering
    \includegraphics[width=\linewidth]{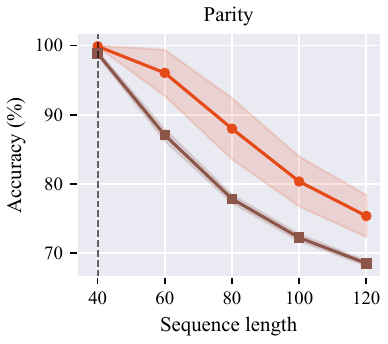}
    \caption{Parity}
    \label{fig:noconv_parity}
\end{subfigure}
\hfill
\begin{subfigure}[t]{0.32\linewidth}
    \centering
    \includegraphics[width=\linewidth]{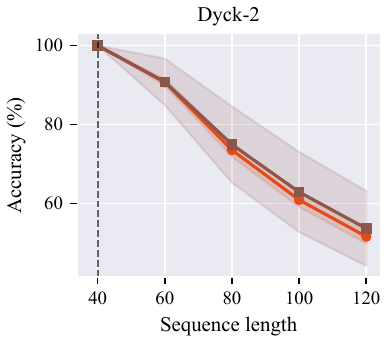}
    \caption{Dyck-2}
    \label{fig:noconv_dyck}
\end{subfigure}
\hfill
\begin{subfigure}[t]{0.32\linewidth}
    \centering
    \includegraphics[width=\linewidth]{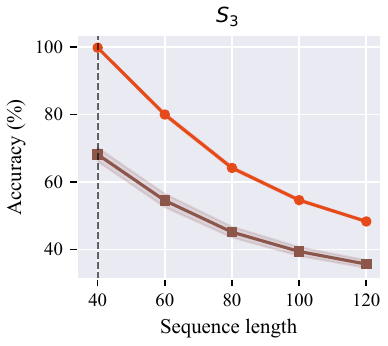}
    \caption{$S_3$}
    \label{fig:noconv_s3}
\end{subfigure}

\vspace{0.3em}
\centerline{\includegraphics[width=0.75\linewidth]{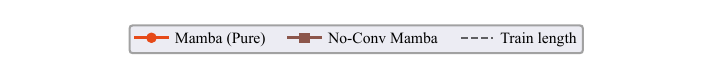}}
\caption{Conv1D ablation under a matched pure-PyTorch backend, so that removing the convolution is not confounded with switching kernel implementation. $n{=}4$ seeds per curve on Parity and $S_3$; on Dyck-2, $n{=}8$ for Mamba (Pure) and $n{=}4$ for No-Conv Mamba.}
\label{fig:noconv}
\end{figure}

\input{tables/tab2_summary}

\section{Additional controls and ablations}
\label{app:extra}

\begin{figure}[!t]
\centering
\begin{subfigure}[t]{0.48\linewidth}
    \centering
    \includegraphics[width=\linewidth]{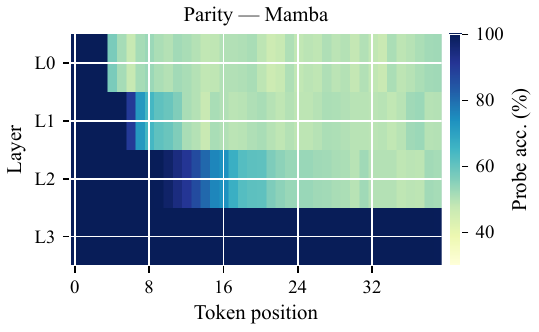}
    \caption{Parity -- Mamba}
\end{subfigure}
\hfill
\begin{subfigure}[t]{0.48\linewidth}
    \centering
    \includegraphics[width=\linewidth]{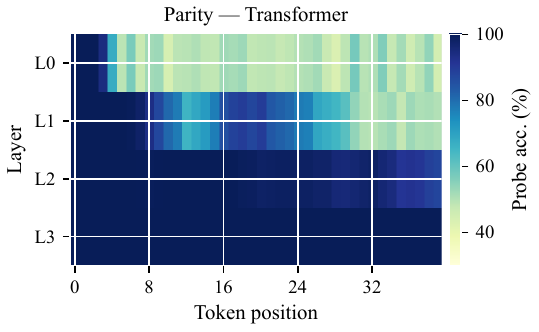}
    \caption{Parity -- Transformer}
\end{subfigure}

\vspace{0.3em}
\begin{subfigure}[t]{0.48\linewidth}
    \centering
    \includegraphics[width=\linewidth]{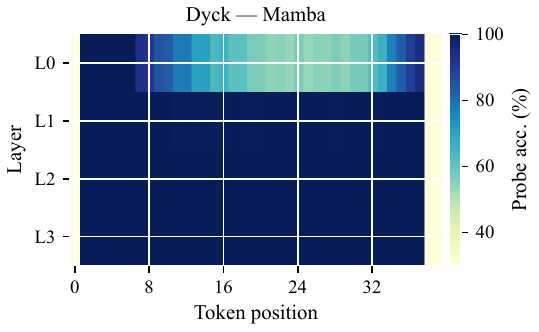}
    \caption{Dyck -- Mamba}
\end{subfigure}
\hfill
\begin{subfigure}[t]{0.48\linewidth}
    \centering
    \includegraphics[width=\linewidth]{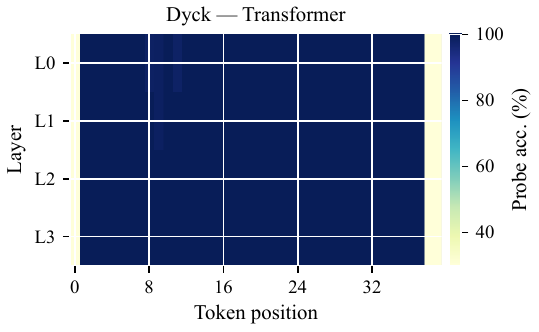}
    \caption{Dyck -- Transformer}
\end{subfigure}
\caption{Per-layer probe accuracy, single-seed snapshot for visual clarity. Chance-normalized $n{=}4$ aggregates for all five architectures on all three tasks are in Appendix~\ref{app:extra}, Table~\ref{tab:norm_probe}.}
\label{fig:probes}
\end{figure}

Chance-normalized probing answers a comparability objection. The three formal tasks differ in output-class count (Parity: 2, $S_3$: 6, Dyck-2: 11), so raw per-layer probe accuracy is not directly comparable across tasks and leaves open the objection that the Fig.~\ref{fig:probes} reversal is a probe-capacity $\times$ class-count artifact rather than a representational reorganization. We rerun the probes under an excess-over-chance normalization, $\tilde a = (a - c)/(1 - c)$ with $c \in \{1/2, 1/6, 1/11\}$, on all three tasks, all five architectures, and four fine-tuning seeds per cell. Table~\ref{tab:norm_probe} reports the result.

\begin{table*}[ht]
\centering
\footnotesize
\setlength{\tabcolsep}{3pt}
\caption{Chance-normalized per-layer probe accuracy, $\tilde a = (a-c)/(1-c)$, mean $\pm$ std across $n{=}4$ fine-tuning seeds. Parity chance $c{=}1/2$; $S_3$ chance $c{=}1/6$; Dyck-2 chance $c{=}1/11$.}
\label{tab:norm_probe}
\begin{tabular}{@{}llccccc@{}}
\toprule
\textbf{Task} & \textbf{Model} & \textbf{$n$} & \textbf{L0} & \textbf{L1} & \textbf{L2} & \textbf{L3} \\
\midrule
Parity & Mamba       & 4 & $0.107{\scriptstyle\pm0.007}$ & $0.211{\scriptstyle\pm0.046}$ & $0.454{\scriptstyle\pm0.099}$ & $0.997{\scriptstyle\pm0.004}$ \\
Parity & Mamba-2     & 4 & $0.114{\scriptstyle\pm0.012}$ & $0.229{\scriptstyle\pm0.044}$ & $0.431{\scriptstyle\pm0.081}$ & $0.999{\scriptstyle\pm0.001}$ \\
Parity & Transformer & 4 & $0.106{\scriptstyle\pm0.012}$ & $0.394{\scriptstyle\pm0.175}$ & $0.659{\scriptstyle\pm0.288}$ & $0.999{\scriptstyle\pm0.003}$ \\
Parity & LSTM        & 4 & $0.100{\scriptstyle\pm0.010}$ & $0.140{\scriptstyle\pm0.009}$ & $0.203{\scriptstyle\pm0.018}$ & $1.000{\scriptstyle\pm0.000}$ \\
Parity & GRU         & 4 & $0.428{\scriptstyle\pm0.381}$ & $0.516{\scriptstyle\pm0.333}$ & $0.819{\scriptstyle\pm0.362}$ & $1.000{\scriptstyle\pm0.000}$ \\
\midrule
$S_3$  & Mamba       & 4 & $0.166{\scriptstyle\pm0.003}$ & $0.303{\scriptstyle\pm0.013}$ & $0.555{\scriptstyle\pm0.032}$ & $0.996{\scriptstyle\pm0.001}$ \\
$S_3$  & Mamba-2     & 4 & $0.178{\scriptstyle\pm0.004}$ & $0.275{\scriptstyle\pm0.017}$ & $0.442{\scriptstyle\pm0.027}$ & $0.948{\scriptstyle\pm0.018}$ \\
$S_3$  & Transformer & 4 & $0.181{\scriptstyle\pm0.011}$ & $0.285{\scriptstyle\pm0.015}$ & $0.455{\scriptstyle\pm0.049}$ & $0.913{\scriptstyle\pm0.040}$ \\
$S_3$  & LSTM        & 4 & $0.139{\scriptstyle\pm0.016}$ & $0.192{\scriptstyle\pm0.088}$ & $0.263{\scriptstyle\pm0.079}$ & $1.000{\scriptstyle\pm0.000}$ \\
$S_3$  & GRU         & 4 & $0.243{\scriptstyle\pm0.100}$ & $0.308{\scriptstyle\pm0.084}$ & $0.841{\scriptstyle\pm0.319}$ & $1.000{\scriptstyle\pm0.000}$ \\
\midrule
Dyck-2 & Mamba       & 4 & $0.811{\scriptstyle\pm0.093}$ & $0.999{\scriptstyle\pm0.001}$ & $1.000{\scriptstyle\pm0.000}$ & $1.000{\scriptstyle\pm0.000}$ \\
Dyck-2 & Mamba-2     & 4 & $0.829{\scriptstyle\pm0.069}$ & $0.999{\scriptstyle\pm0.001}$ & $1.000{\scriptstyle\pm0.000}$ & $1.000{\scriptstyle\pm0.000}$ \\
Dyck-2 & Transformer & 4 & $0.993{\scriptstyle\pm0.004}$ & $0.999{\scriptstyle\pm0.000}$ & $1.000{\scriptstyle\pm0.000}$ & $1.000{\scriptstyle\pm0.000}$ \\
Dyck-2 & LSTM        & 4 & $0.732{\scriptstyle\pm0.066}$ & $0.998{\scriptstyle\pm0.004}$ & $1.000{\scriptstyle\pm0.000}$ & $1.000{\scriptstyle\pm0.000}$ \\
Dyck-2 & GRU         & 4 & $0.942{\scriptstyle\pm0.063}$ & $1.000{\scriptstyle\pm0.000}$ & $1.000{\scriptstyle\pm0.000}$ & $1.000{\scriptstyle\pm0.000}$ \\
\bottomrule
\end{tabular}
\end{table*}

The reversal survives normalization as a mean effect, with two nuances worth flagging at 4-layer scale. On Parity, LSTM and Mamba remain consistently late-concentrated ($\tilde a_{L_2} = 0.203 \pm 0.018$ and $0.454 \pm 0.099$), while Transformer ($0.659 \pm 0.288$) shows higher mid-layer readability with substantial across-seed variance: per-seed values range from $0.266$ to $0.951$, so individual Transformer seeds lie anywhere between ``distributed buildup across depth'' and ``late concentration.'' GRU Parity is a more extreme outlier: the mean $0.428 \pm 0.381$ at L0 is driven by a single seed whose probe reaches $1.0$ already at the input layer (the gate appears to compute parity directly from the embedding), while the other three seeds sit between $0.21$ and $0.25$. $S_3$ reproduces the Parity pattern for Mamba, Mamba-2, LSTM, and Transformer; GRU is again an early-concentration outlier at $0.841$ at L2. On Dyck-2, Transformer is at ceiling already at L0 ($0.993 \pm 0.004$), while LSTM ($0.732 \pm 0.066$), Mamba ($0.811 \pm 0.093$), and Mamba-2 ($0.829 \pm 0.069$) start below ceiling at L0 and reach it by L1--L2; GRU Dyck ($0.942 \pm 0.063$) is closer to Transformer than to LSTM, so the recurrent/attention dichotomy alone does not determine the profile.

The qualitative reversal (recurrent/SSM models concentrate late on prefix-update tasks, Transformer concentrates early on Dyck) holds as a mean effect across chance-normalized probes at $n{=}4$ seeds per cell, and is not an artifact of the different class counts. At 4-layer scale the gradual-buildup side of the story is compressed into one or two layers; the 24-layer pretrained results in Fig.~\ref{fig:pretrained_probe_compare} unfold it more fully.

\paragraph{Post-attention LayerNorm~\citep{li2025vanishing}.} Adding LayerNorm after the attention output (before the residual) gives $33.82\pm 1.03\%$ on Dyck at $L=120$, which is $3.55$\,pp \emph{below} vanilla Transformer ($37.37\pm 1.34\%$). On Parity only 2 of 4 seeds converge. Attention-side modifications alone do not close the recurrence gap.

\begin{figure}[!tbp]
\centering
\begin{subfigure}[t]{\linewidth}
    \centering
    \includegraphics[width=\linewidth]{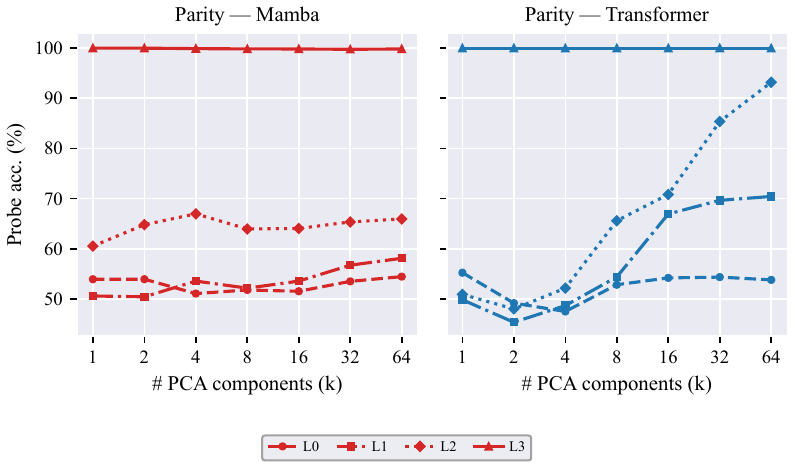}
    \caption{Parity}
\end{subfigure}

\vspace{0.6em}
\begin{subfigure}[t]{\linewidth}
    \centering
    \includegraphics[width=\linewidth]{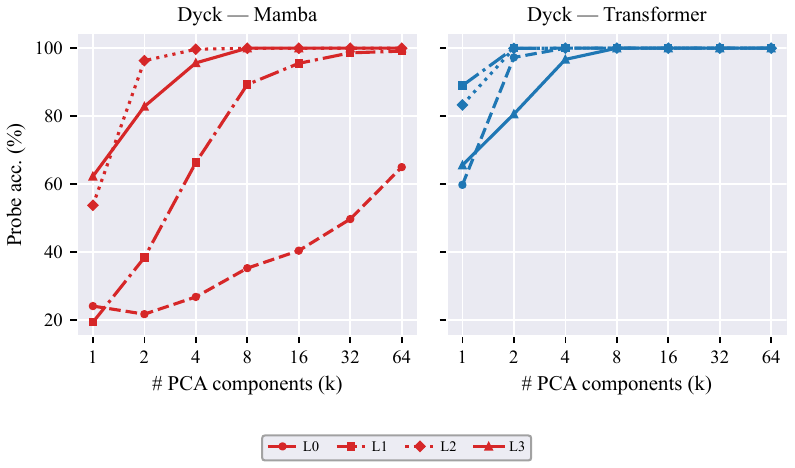}
    \caption{Dyck}
\end{subfigure}
\caption{PCA probe accuracy against number of principal components, single-seed snapshot at the key layer of each (task, architecture) pair. On Parity both models reach near-perfect accuracy with very few components; Dyck needs broader subspaces, especially in early Mamba layers.}
\label{fig:pca}
\end{figure}

\paragraph{PCA of hidden states.} On Parity, a small number of principal components explain most of the final-layer variance for both Mamba and Transformer. On Dyck, early recurrent and Mamba layers occupy visibly broader subspaces, consistent with the slower hierarchical buildup seen in probing. Figure~\ref{fig:pca} shows probe accuracy as a function of the number of principal components used as features.

Training dynamics show the same task dependence. On Parity, Mamba Layer~3 crystallizes abruptly from ${<}60\%$ to ${>}99\%$ in about $1{,}500$ steps, while Transformer layers improve gradually over many thousands of steps. On Dyck the pattern reverses: Transformer reaches near-perfect probe accuracy from the first checkpoint (step~195), while Mamba exhibits hierarchical layer-by-layer emergence. The reversal of probe profiles is therefore not an end-of-training artifact; it is already visible in how each architecture organizes state over the course of training.

\begin{figure}[!tbp]
\centering
\begin{subfigure}[t]{\linewidth}
    \centering
    \includegraphics[width=\linewidth]{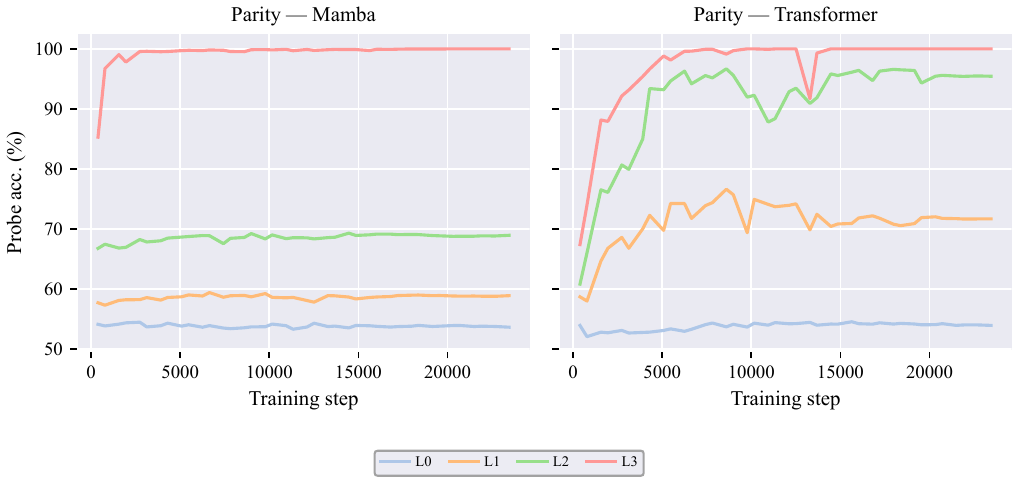}
    \caption{Parity}
\end{subfigure}

\vspace{0.6em}
\begin{subfigure}[t]{\linewidth}
    \centering
    \includegraphics[width=\linewidth]{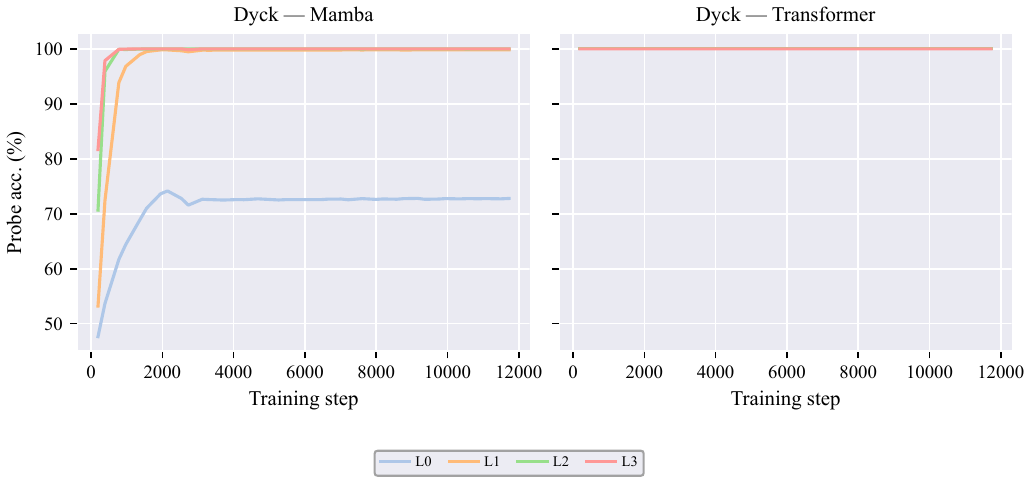}
    \caption{Dyck}
\end{subfigure}
\caption{Per-layer probe accuracy over training, single seed (42). On Parity, Mamba's final layer crystallizes abruptly while Transformer improves gradually; on Dyck, Transformer becomes readable almost immediately while Mamba shows layer-by-layer emergence.}
\label{fig:dynamics}
\end{figure}

At the Dyck-$k$ boundary the gap is flat. At $L=120$, the Mamba--Transformer gap on Dyck is stable across bracket-type complexity: $+15.01$\,pp at $k=2$, $+14.75$\,pp at $k=3$, $+15.17$\,pp at $k=4$. Mamba's accuracy varies by only $0.75$\,pp across $k$, from $51.74\%$ to $52.49\%$, well within the cross-seed noise floor of $\sigma\approx 1$\,pp. The generalization bottleneck is sequence length, not bracket-type complexity.

\begin{figure}[!tbp]
\centering
\includegraphics[width=0.70\linewidth]{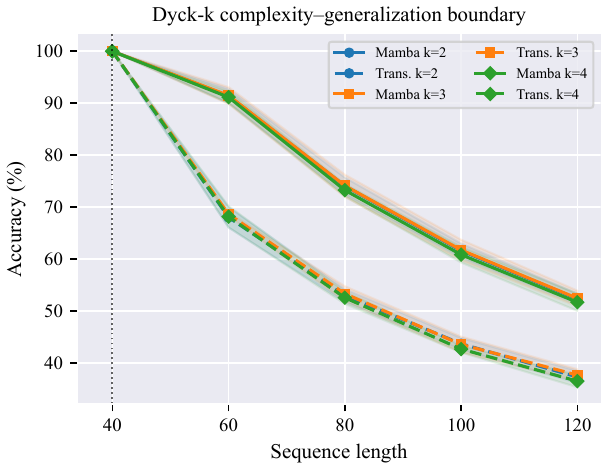}
\caption{Dyck-$k$ generalization boundary ($n=4$ seeds). Mamba's advantage over Transformer is stable across $k=2,3,4$, indicating that the bottleneck is sequence length rather than bracket-type complexity.}
\label{fig:dyck_k}
\end{figure}

\FloatBarrier

\section{Computational environment}
\label{app:compute}

All experiments run on single NVIDIA RTX~3090 GPUs (24\,GB); the Pythia-410M Dyck layer sweep uses four GPUs in parallel, one per fine-tuning seed. Software: Python~3.10, PyTorch~2.4.0 (CUDA~11.8), \texttt{mamba-ssm}~2.2.1, \texttt{causal-conv1d}~1.5.0, HuggingFace Transformers~4.39, scikit-learn~1.7.2. The dominant analyses by wall-clock are the pure-PyTorch Conv1D ablation (no fused kernel available for $d_\text{conv}=1$) and the layer-sweep causal interventions on pretrained models, with the Pythia-410M Dyck layer sweep the most expensive single component.

\input{sections/appendix_controls}

%% file: tables/tab1_config.tex
\begin{table*}[ht]
\caption{Model configurations for the $4$-layer formal-task experiments. All models use $d_\text{model}=128$ unless noted; matched variants increase $d_\text{model}$ to match $\sim\!830$K parameters. Non-embedding parameter counts.}
\label{tab:config}
\centering
\footnotesize
\setlength{\tabcolsep}{4pt}
\begin{tabular}{@{}llll@{}}
\toprule
\textbf{Model} & \textbf{Architecture} & \textbf{Key config} & \textbf{Params} \\
\midrule
Transformer      & $4$-head attn $+$ FFN    & $d_\text{ff}=512$, learned PE              & ${\sim}827$K \\
Trans.$+$ALiBi   & Same, ALiBi PE           & No learned positions                       & ${\sim}794$K \\
Trans.$+$RoPE    & Same, RoPE               & $\theta=10000$                             & ${\sim}794$K \\
Gated Trans.     & Attn $+$ FFN $+$ GRU gate & Per-layer gate after attn                 & ${\sim}827$K \\
Mamba (Full)     & Conv1D $+$ Sel.\ SSM     & $d_\text{state}=44$, $d_\text{conv}=4$     & ${\sim}830$K$^\dagger$ \\
Mamba-2 (SSD)    & Multi-head SSM           & $d_\text{state}=24$, exp$=4$, hd$=64$      & ${\sim}830$K$^\dagger$ \\
Mamba (Pure)     & Same, pure PyTorch       & No CUDA kernel                             & ${\sim}879$K \\
Mamba (No-Conv)  & SSM only (pure)          & $d_\text{conv}=1$                          & ${\sim}874$K \\
LSTM             & $4\times$ LSTM cell      & Pre-LN, residual                           & ${\sim}530$K \\
GRU              & $4\times$ GRU cell       & Pre-LN, residual                           & ${\sim}398$K \\
LSTM (matched)   & $4\times$ LSTM cell      & $d=160$                                    & ${\sim}827$K \\
GRU (matched)    & $4\times$ GRU cell       & $d=185$                                    & ${\sim}828$K \\
\bottomrule
\end{tabular}

\vspace{0.3em}
\raggedright\scriptsize
$^\dagger$Official \texttt{mamba-ssm} CUDA kernel. The ${\sim}49$K gap between Full and Pure arises from re-parameterization of the fused kernel; the Conv1D ablation uses pure-PyTorch for both conditions ($\Delta$params $=4{,}608$).
\end{table*}

%% file: tables/tab_inlp_pretrained.tex
\begin{table*}[ht]
\centering
\footnotesize
\caption{Subspace erasure at pretrained scale, at each cell's first readable layer, with a budget of
$256$ directions rather than the $100$ used at $128$ channels. $^{\dagger}$ marks cells that reached
the budget before a probe fell to chance.}
\label{tab:inlp_pretrained}
\setlength{\tabcolsep}{3pt}
\begin{tabular}{@{}llccccccc@{}}
\toprule
& & & & \multicolumn{2}{c}{\textbf{Probe}} & \multicolumn{2}{c}{\textbf{Relative drop}} & \\
\cmidrule(lr){5-6}\cmidrule(lr){7-8}
\textbf{Task} & \textbf{Model} & \textbf{Layer} & $k$ & \textbf{before} & \textbf{after} & \textbf{probe} & \textbf{random} & $n$ \\
\midrule
Parity & Pythia-160M & L9--L11 & $140\pm 78$$^{\dagger}$ & $0.981\pm 0.003$ & $0.538\pm 0.041$ & $0.713\pm 0.352$ & $0.024\pm 0.015$ & $4$ \\
Parity & Mamba-130M & L22--L23 & $255\pm 2$$^{\dagger}$ & $0.980\pm 0.014$ & $0.635\pm 0.095$ & $0.886\pm 0.191$ & $0.013\pm 0.012$ & $4$ \\
Dyck-2 & Pythia-160M & L6--L10 & $256\pm 0$$^{\dagger}$ & $0.985\pm 0.007$ & $0.420\pm 0.046$ & $0.918\pm 0.100$ & $0.418\pm 0.149$ & $4$ \\
Dyck-2 & Mamba-130M & L20--L23 & $256\pm 0$$^{\dagger}$ & $0.987\pm 0.012$ & $0.492\pm 0.197$ & $0.713\pm 0.149$ & $0.103\pm 0.091$ & $4$ \\
\bottomrule
\end{tabular}
\end{table*}

%% file: tables/tab_pretrained_causal.tex
\begin{table*}[ht]
\centering
\footnotesize
\caption{Pretrained single-direction zero-ablation under the unified residual tap, mean $\pm$ std
over $n{=}4$ fine-tuning seeds. Probe and ablation act on the same tensor $h_i$, the residual
stream after block $i$. Drops are relative, (clean $-$ ablated) over (clean $-$ chance);
``$\times$ null'' compares the ablation peak against random unit directions at that layer, and a
dash marks a cell that does not reach three times it. Ablation depth is that layer over
$L_{\max}$.}
\label{tab:pretrained_causal}
\setlength{\tabcolsep}{4pt}
\begin{tabular}{@{}llccccccc@{}}
\toprule
& & \multicolumn{3}{c}{\textbf{At the probe peak}} & \multicolumn{4}{c}{\textbf{At the ablation peak}} \\
\cmidrule(lr){3-5}\cmidrule(lr){6-9}
\textbf{Model} & \textbf{Task} & \textbf{Layer} & \textbf{Probe acc} & \textbf{Drop} & \textbf{Layer} & \textbf{Drop} & \textbf{$\times$ null} & \textbf{Ablation depth} \\
\midrule
Pythia-160M & Dyck-2 & L11 & $98.8\%$ & $18.63\pm 3.85$ & L7 & $30.24\pm 19.00$ & $83\times$ & $0.64$ \\
Pythia-160M & Parity & L11 & $97.4\%$ & $3.60\pm 6.36$ & L11 & $3.60\pm 6.36$ & $60\times$ & $1.00$ \\
Pythia-160M & $S_3$ & L11 & $63.0\%$ & $1.67\pm 0.60$ & L11 & $1.67\pm 0.60$ & $8\times$ & $1.00$ \\
Pythia-410M & Dyck-2 & L22 & $98.8\%$ & $3.79\pm 4.48$ & L9 & $21.22\pm 22.95$ & $41\times$ & $0.39$ \\
Mamba-130M & Dyck-2 & L23 & $99.0\%$ & $0.67\pm 0.64$ & L19 & $46.29\pm 19.85$ & $141\times$ & $0.83$ \\
Mamba-130M & Parity & L23 & $97.9\%$ & $0.02\pm 0.02$ & -- & -- & -- & -- \\
Mamba-130M & $S_3$ & L23 & $78.7\%$ & $0.06\pm 0.03$ & -- & -- & -- & -- \\
\bottomrule
\end{tabular}
\end{table*}

%% file: tables/tab_rowspace_specificity.tex
\begin{table*}[ht]
\centering
\footnotesize
\caption{Ablating the probe's full $K{-}1$ row space against random subspaces of the \emph{same}
dimension, $n{=}4$ seeds. ``First specific layer'' is the shallowest layer whose row-space drop
reaches three times its matched random null; ``$\le$ random'' counts layers where the probe's own
directions do no more damage than the same number of arbitrary ones.}
\label{tab:rowspace_specificity}
\setlength{\tabcolsep}{4pt}
\begin{tabular}{@{}llccccc@{}}
\toprule
\textbf{Model} & \textbf{Task} & $K{-}1$ & \textbf{First specific layer} & \textbf{$\le$ random} & \textbf{Max drop} & \textbf{Max null} \\
\midrule
Pythia-160M & Dyck-2 & $10$ & L1 & $0/12$ & $46.4\%$ & $12.9\%$ \\
Pythia-160M & Parity & $1$ & L10 & $10/12$ & $3.6\%$ & $0.5\%$ \\
Pythia-160M & $S_3$ & $5$ & L10 & $10/12$ & $4.2\%$ & $2.4\%$ \\
Pythia-410M & Dyck-2 & $10$ & L2 & $2/24$ & $35.8\%$ & $9.4\%$ \\
Mamba-130M & Dyck-2 & $10$ & L13 & $5/24$ & $54.0\%$ & $16.3\%$ \\
Mamba-130M & Parity & $1$ & L21 & $22/24$ & $0.2\%$ & $2.4\%$ \\
Mamba-130M & $S_3$ & $5$ & L21 & $20/24$ & $2.1\%$ & $3.6\%$ \\
\bottomrule
\end{tabular}
\end{table*}

%% file: tables/tab2_summary.tex
\begin{table*}[t]
\caption{The four measurement axes, across architectures and tasks. The Dyck column is Dyck-2 with $d_\text{max}=10$, whose label is nesting depth, that is stack size.}
\label{tab:summary}
\centering
\footnotesize
\setlength{\tabcolsep}{3pt}
\begin{tabular}{@{}lccc@{}}
\toprule
\textbf{Measure} & \textbf{Parity} & \textbf{$\boldsymbol{S_3}$} & \textbf{Dyck-2} \\
\midrule
\multicolumn{4}{@{}l}{\emph{Length-generalization accuracy at $L=120$}} \\
\quad LSTM         & $100.0 \pm 0.0\%$            & $100.0 \pm 0.0\%$            & $61.6 \pm 4.8\%$ \\
\quad Mamba        & $\hphantom{0}77.3 \pm 2.1\%$ & $\hphantom{0}48.2 \pm 0.7\%$ & $52.4 \pm 1.2\%$ \\
\quad Transformer  & $\hphantom{0}66.6 \pm 0.3\%$ & $\hphantom{0}42.1 \pm 1.2\%$ & $37.4 \pm 1.3\%$ \\
\quad\emph{Ordering} & \multicolumn{3}{c}{\textbf{Recurrent $>$ SSM $>$ Transformer} (all three tasks)} \\
\midrule
\multicolumn{4}{@{}l}{\emph{Subspace erasure at the readable layer}, relative drop (random subspace of the same size)} \\
\quad Transformer  & $94.5 \pm \hphantom{0}5.3\%$ ($0.1$) & $96.8 \pm 1.0\%$ ($3.9$) & $93.2 \pm \hphantom{0}4.0\%$ ($58.6$) \\
\quad Mamba        & $46.4 \pm 39.6\%$ ($0.0$) & $97.4 \pm 3.0\%$ ($1.0$) & $69.9 \pm 16.1\%$ ($30.7$) \\
\quad Mamba-2      & $97.2 \pm \hphantom{0}3.3\%$ ($0.0$) & $98.1 \pm 1.1\%$ ($2.1$) & $85.0 \pm \hphantom{0}2.5\%$ ($35.4$) \\
\midrule
\multicolumn{4}{@{}l}{\emph{Layerwise probe profile}: first layer within $0.5$\,pp of that cell's best, 4-layer models} \\
\quad Recurrent (LSTM/GRU) & L3 & L3 & L1 \\
\quad SSM (Mamba/Mamba-2)  & L3 & L3 & L1--L2 \\
\quad Transformer          & L3 & L3 & \textbf{L0} \\
\midrule
\multicolumn{4}{@{}l}{\emph{Conv1D ablation effect on Mamba} (accuracy gap, NoConv $-$ Full Pure)} \\
\quad Accuracy gap & $\mathbf{-6.83\,\text{pp}}$ & $\mathbf{-12.64\,\text{pp}}$ & $+4.04\,\text{pp}$ (n.s.) \\
\bottomrule
\end{tabular}

\vspace{0.4em}
\raggedright\scriptsize
Mean~$\pm$~std across $n=4$ seeds, except Conv1D on Dyck (Mamba Pure $n=8$ against No-Conv $n=4$). Relative drop is (clean $-$ ablated) over (clean $-$ chance); the parenthesized figure deletes a random subspace of the same dimension at the same layer, and on Dyck that dimension is most of the layer width, which is why the two are close there. Recurrent rows and the full table, including the shared-budget comparison, are in Table~\ref{tab:inlp}. Accuracy gap is NoConv $-$ Full Pure Mamba, so \emph{negative} values mean Conv1D helps; n.s.\ is $p > 0.2$ (Welch $t$-test). Conv1D is a Mamba-specific component, hence the single row. Chance-normalized probe aggregates are in Table~\ref{tab:norm_probe}.
\end{table*}

%% file: sections/appendix_controls.tex
\FloatBarrier

\section{Controls and scope tests}
\label{app:controls}

This appendix carries the controls the main text relies on, together with the tests that fix how far
the picture extends.

Each scope test has a pass/fail rule fixed in writing before any model for it was trained, applied
mechanically afterwards and not revised once results were visible. Six were run, and each is reported
below against the rule it was judged by.

\subsection{Changing only the supervised target changes the profile}
\label{app:targetswap}

The sharpest control we can offer holds everything fixed except what the model is trained to predict. On
the same Dyck-2 inputs we train one group of models to predict the type of the exposed top of stack
(stack \emph{content}) and another to predict the current nesting depth (stack \emph{size}). We then
probe both groups for the same variable, the exposed top of stack, at the same positions.

\paragraph{Shortcut control.} We score only \emph{query positions}: a closing bracket that pops the stack,
leaves it non-empty, and exposes an item opened at least eight tokens earlier. At these positions a probe
on the current token alone reaches $.4956 \pm .0045$ and a position-only probe $.4985 \pm .0028$, so
neither reveals the answer. A bag-of-counts probe does reach $.7547 \pm .0058$, so we additionally
evaluate on a contrast set matched on the full prefix-count vector and on the last four tokens, with the
sequence-level train/test split fixed before matching. On that set the count baseline, the last-four
baseline, and their combination are all exactly $0.500$ by construction, verified as
$\min = \max = 0.5000$ over all cells.

Table~\ref{tab:targetswap} gives balanced accuracy on the contrast set.

\begin{table*}[t]
\centering
\small
\caption{Probe accuracy for the exposed top of stack, at the same query positions and on the same
contrast set, for models that differ only in what they were trained to predict (four model seeds).
Layer 2 is the comparison of interest, before the supervised readout. The binary-depth rows match the
content rows on the number of output classes.}
\label{tab:targetswap}
\begin{tabular}{llcccc}
\toprule
trained to predict & arch & L0 & L1 & L2 & L3 \\
\midrule
stack content (top type)  & Transformer & $.530$ & $.728$ & $\mathbf{.961 \pm .058}$ & $1.000$ \\
stack content (top type)  & Mamba       & $.573$ & $.630$ & $\mathbf{.804 \pm .037}$ & $1.000$ \\
stack size (exact depth)  & Transformer & $.498$ & $.545$ & $.543$ & $.542$ \\
stack size (exact depth)  & Mamba       & $.500$ & $.504$ & $.503$ & $.496$ \\
stack size (binary depth) & Transformer & $.500$ & $.547$ & $\mathbf{.560 \pm .011}$ & $.572$ \\
stack size (binary depth) & Mamba       & $.535$ & $.529$ & $\mathbf{.548 \pm .020}$ & $.525$ \\
\bottomrule
\end{tabular}
\end{table*}

The key comparison is L2, before the supervised readout. The binary-depth rows are an
output-cardinality-matched control, so the content and size groups are matched on the number of output
classes as well as on inputs and probed positions. The Transformer content cell carries the largest
spread in the table and should not be read as tight.

A second target, the next valid closing bracket, is at a fixed position a
relabeling of the first, but it is defined at every position where the stack is non-empty rather than
only at the probed positions, which removes any concern that we supervise exactly where we probe.
Probing the same query positions on the same contrast set gives Transformer $.532 / .989 / 1.000 /
1.000$ and Mamba $.568 / .634 / .855 / 1.000$. The Transformer L1 value moves from $.728$ to $.989$
while Mamba's L2 moves only from $.804$ to $.855$.

\subsection{Controls on the probe itself}

\paragraph{Selectivity} (following \citealp{hewitt2019designing}). For each task, architecture and layer we train a
control probe on labels whose whole trajectories are shuffled across sequences, with matched capacity and
budget, using ten permutations across three model seeds. Control accuracy stays at the
position-conditional baseline with no layer structure: baselines are $.5000$ (Parity), $.2423$ (Dyck) and
$.1771$ ($S_3$), and control means stay within $[.4980, .5020]$, $[.2415, .2432]$ and $[.1747, .1788]$
respectively. Real probes reach $1.000$ on Dyck and $.9997$ to $1.000$ on Parity. The one cell that does
not approach $1.0$ is $S_3$ on Transformer, at $.9383$.

\paragraph{Nonlinear probe.} Repeating the target-swap analysis with the one-hidden-layer probe used
elsewhere in this paper does not change the picture. On content-supervised models the MLP agrees with the
linear probe and is slightly higher (Transformer L2 $.961$ linear against $.969$ MLP; Mamba L2 $.804$
against $.852$). On size-supervised models it recovers nothing extra and is generally lower (binary
depth, Transformer L2 $.560$ against $.525$, L3 $.572$ against $.517$). The dissociation is therefore not
an artifact of restricting the probe to be linear.

\subsection{Matched output cardinality and matched task accuracy}

To separate task difficulty from the task contrast we ran a nine-cell grid with architecture, model size,
training budget, sequence length and probe protocol fixed, sweeping modular prefix updates at state sizes
$2,3,4,5,6,24$ against stack-content targets with $2,3,4$ bracket types. The metric is a chance-corrected
profile AUC, $(\text{raw} - \text{chance}) / (1 - \text{chance})$, since raw accuracy has a
class-count-dependent floor.

On Mamba, restricted to cells reaching at least $0.99$ task accuracy, the three cardinality-matched pairs
separate: at $2$ classes $.326$ against $.689$, at $3$ classes $.236$ against $.696$, at $4$ classes
$.242$ against $.717$. Task accuracies are $.9992$, $.9993$ and $.9950$ on the prefix-update side and
$1.0000$ throughout on the stack-content side. Three qualifications matter. Extending the prefix-update
sweep gives $.382$ at state size $5$, so the profile does not order by state size. $S_3$ and $S_4$ reach
only $.8822$ and $.7741$ task accuracy under the shared budget and are secondary evidence at best.
Finally, our Transformer runs on the modular cells did not reach a comparable accuracy under this budget,
so we present the accuracy-matched comparison for Mamba only; a separate budget diagnostic, rather than
this grid, is what establishes that point.

\subsection{Frozen probing at 1.4B, and what does not replicate}

We probed frozen Pythia-1.4B and Mamba-1.4B without updating either backbone, rendering the formal tasks
in each model's own vocabulary. Layer indexing is aligned across the two families before comparison:
HuggingFace Mamba appends a hidden state after each block while GPT-NeoX records one before each block,
so an index-by-index comparison is off by one. We detect the embedding row by measurement rather than by
name, and confirm the alignment by checking that the first row of the top-of-stack probe sits at chance
($.4962$).

With two seeds, which is weaker than the four used elsewhere in this paper, stack depth peaks at $.7323$
at layer $3$ of $24$ in Pythia and at $.7637$ at layer $24$ of $49$ in Mamba (chance $.0909$). Top of
stack at query positions peaks at $.8586$ at layer $7$ of $24$ in Pythia and $.8441$ at layer $27$ of
$49$ in Mamba (chance $.5$). The Dyck-side early-versus-gradual pattern therefore does appear at this
scale. Parity and $S_3$ stay at or barely above chance in both models (Parity peaks $.5523$ and $.5618$;
$S_3$ peaks $.2170$ and $.2783$), so this setting cannot test the other side of the contrast at all.

Separately, the four-layer content-versus-size dissociation of Appendix~\ref{app:targetswap} does
\emph{not} reproduce at $1.4$B. A first pass suggested an architecture effect, with one Pythia run near
chance and one Mamba run at $.808$, but those runs differed in task accuracy as well as architecture. We
therefore probed each model during training at the first point it reached a fixed accuracy target, and
again at the end. At matched accuracy all four checkpoints agree: Pythia $.8436$ and $.8556$ at task
accuracy $.9863$ and $.9894$, Mamba $.8485$ and $.8710$ at task accuracy $1.0000$. We restrict the
dissociation claim to the controlled four-layer setting. The first pass is worth reporting as a caution
in its own right: a difference in training state can look exactly like an architecture effect, and
probing at matched accuracy is what distinguishes them.

\subsection{Scope tests}

Each of the following fixed its rule in advance. None is a post-hoc reinterpretation.

\paragraph{1. Predictive power on unseen task families.} We specified in advance which side of the contrast
three held-out designs should fall on and what band would count as success, then trained them under the
same architecture, budget and probe protocol. Chance-corrected AUC, Transformer and Mamba: carry
arithmetic in base 4, predicted below $0.50$, gives $.9412$ and $.9955$; three counters modulo 3,
predicted below $0.50$, gives $.5174$ and $.3705$; the element one below the top of a Dyck-2 stack,
predicted above $0.60$, gives $.6543$ and $.5852$. Counting a prediction as met only when it holds for
both architectures, the score is $0$ of $3$. The carry task misses widely, and a current-column probe
already reaches $.876$ on it, so a locally predictable label scores high regardless of family. We drop
the taxonomy and the predictive claim.

\paragraph{2. Refinement order.} Writing $T_1$ for the fine top-of-stack type (4 classes) and $T_2$ for a
coarse 2-class relabeling of it, we predicted that $T_1$-trained models would expose $T_2$ while
$T_2$-trained models would not expose $T_1$, since a coarse objective gives no gradient pressure to
separate types within a coarse class. The rule required $T_2$-trained models to read $T_1$ at no better
than $0.30$ at every layer. The first half held: $T_1$-trained models read $T_2$ at $1.000$
(Transformer) and $.999$ (Mamba). The second half failed clearly. $T_2$-trained models read $T_1$ up to
$.591$ (Transformer) and $.506$ (Mamba) against a matched-set chance of $.251$. The coarse objective
preserves much of the fine distinction anyway, so the asymmetry is graded rather than categorical and
the rule is not supported.

\paragraph{3. Supervision density.} We wrote down, before running it, the hypothesis that the coverage effect in
Appendix~\ref{app:targetswap} is driven by how \emph{many} positions carry supervision, sweeping the
fraction of retained supervised positions over $.02, .05, .10, .215, .50, 1.0$. One observation is worth
recording about the rule itself: the statistic we had fixed on, a normalized onset layer, turned out to be
\emph{constant} across all six densities in both architectures ($0.333$ for Transformer, $1.000$ for
Mamba), so the rank correlation the rule called for is undefined and we fall back to the per-layer
accuracies. Those refute the hypothesis. The Transformer is flat, reading $.983, .984, .963, .961, .963,
.988$ at layer 1 across the six densities, a range of $.026$. Mamba is the architecture that responds,
reading $.667, .710, .760, .806, .835, .843$ at layer 2, a range of $.176$. Comparing against
Appendix~\ref{app:targetswap}, two percent random supervision gives the Transformer $.983$ at layer 1
while query-only supervision, with far more labels, gives $.728$; and Mamba's query-only value of $.804$
sits essentially on its $p = 0.215$ value of $.806$. What separates the two supervision regimes is
therefore which positions carry supervision, not how many. That reading is post-hoc and would need its
own test, fixed in advance.

\paragraph{4. Attention placement in a hybrid.} A minimal interleaved attention and SSM model showed an
interaction rather than a clean component effect. Two accounts of that interaction, both fixed in advance,
failed. The type of the probed block gives a gap of $+0.122$ against a required $0.15$. The number of
attention blocks is contradicted by within-count spreads of $.486$, $.474$ and $.260$ across placements
that hold the count fixed, against a required maximum of $0.10$.

\paragraph{5. A causal counterpart to the target dissociation.} We asked whether size-supervised models
concentrate task state in a causally fragile direction while content-supervised models do not. They do
not differ. Reporting the relative accuracy drop from ablating the probe direction, taken as each seed's
peak over layers and then averaged over four seeds, the content targets give $.724 \pm .176$ and
$.954 \pm .052$ (Transformer) against $.467 \pm .310$ and $.400 \pm .219$ (Mamba), and the size targets
give $.781 \pm .062$ and $.713 \pm .338$ (Transformer) against $.792 \pm .145$ and $.557 \pm .322$
(Mamba). Every target has a strongly fragile direction and there is no content-versus-size pattern.
Random-direction controls reach at most $0.0008$ across all $128$ cells. The seed spreads are large,
particularly for the binary-depth target, and we would not read fine differences between these cells.

\paragraph{6. Attention placement, confirmatory test.} A regularity noticed after the fact in the hybrid
sweep, that layer-2 readability tracks the type of the top block, replicated cleanly on four fresh seeds
(group means $.9995$ against $.6467$, a gap of $+0.353$, with no overlap between the two groups of seven
placements). It did not extend. At depth 6, where the block immediately above the probe and the top block
are no longer the same block, neither candidate predicted layer-3 readability ($d = -0.011$ against a
$\pm 0.20$ threshold). A third rule suggested by the same data was checked against two independent cells
before being written down and fails on both. We report the replication and the failed extension together
and claim no mechanism.

The per-cell data and the analysis scripts that apply each rule are available from the author, so every verdict above can be recomputed from the recorded
results.